\documentclass[10pt,twocolumn,letterpaper]{article}

\usepackage[pagenumbers]{cvpr} 

\definecolor{cvprblue}{rgb}{0.21,0.49,0.74}
\usepackage[pagebackref,breaklinks,colorlinks,allcolors=cvprblue]{hyperref}

\def\paperID{21165} 
\def\confName{CVPR}
\def\confYear{2026}

\title{What You See is (Usually) What You Get: Multimodal Prototype Networks that Abstain from Expensive Modalities}

\author{Muchang Bahng\thanks{Equal contribution.}\\
Duke University\\
{\tt\small muchang.bahng@duke.edu}
\and
Charlie Berens\footnotemark[1]\\
Duke University\\
{\tt\small cjb131@duke.edu}
\and
Eric Chen\\
Duke University\\
{\tt\small eric.y.chen@duke.edu}
\and
Jon Donnelly\\
Duke University\\
{\tt\small jon.donnelly@duke.edu}
\and
Chaofan Chen\\
University of Maine\\
{\tt\small chaofan.chen@maine.edu}
\and
Cynthia Rudin\\
Duke University\\
{\tt\small cynthia@cs.duke.edu}
}

\begin{document}
\maketitle


\begin{abstract}
  Species detection is important for monitoring the health of ecosystems and identifying invasive species, serving a crucial role in guiding conservation efforts.
  Multimodal neural networks have seen increasing use for identifying species to help automate this task, but they have two major drawbacks. 
  First, their black-box nature prevents the interpretability of their decision making process. Second, collecting genetic data is often expensive and requires invasive procedures, often necessitating researchers to capture or kill the target specimen. 
  We address both of these problems by extending prototype networks (ProtoPNets), which are a popular and interpretable alternative to traditional neural networks, to the multimodal, cost-aware setting. 
  We ensemble prototypes from each modality, using an associated weight to determine how much a given prediction relies on each modality. 
  We further introduce methods to identify cases for which we do not need the expensive genetic information to make confident predictions. 
  We demonstrate that our approach can intelligently allocate expensive genetic data for fine-grained distinctions while using abundant image data for clearer visual classifications and achieving comparable accuracy to models that consistently use both modalities. 
\end{abstract}

\section{Introduction}
\label{sec:intro} 

    \begin{figure}[t]
      \centering
      \includegraphics[width=\columnwidth]{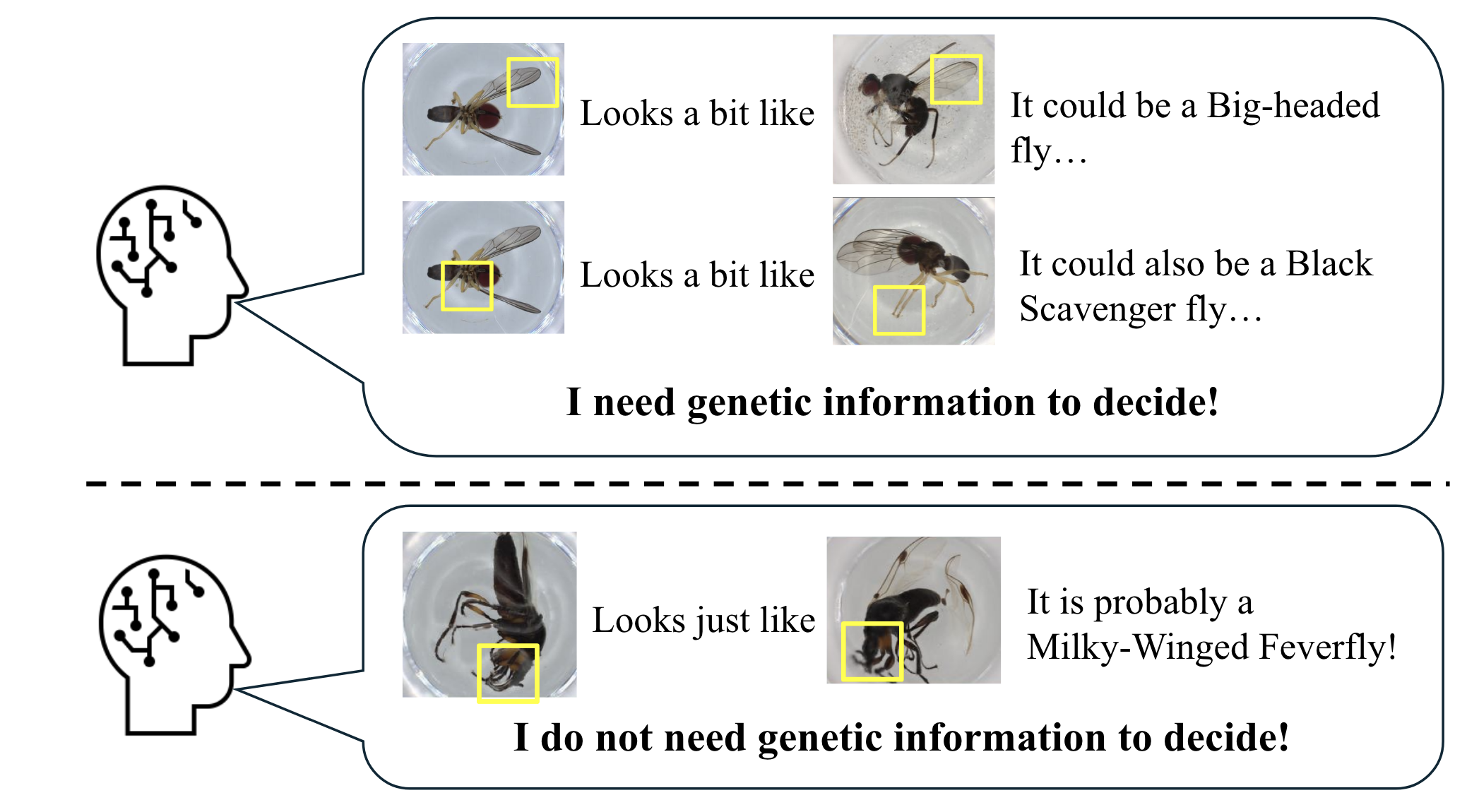}
       \caption{In practice, some samples are easier to classify than others -- if visual information is sufficient to classify a sample, we need not measure genetic information. Our models intelligently decide when they can abstain from measuring genetic information while maintaining accuracy and follow a transparent reasoning process.}
       \label{fig:introfig}
    \end{figure}

    In biological domains such as ecology and healthcare, machine learning tools have been used to help professionals make more informed decisions. These high-stakes problems often use multiple streams of data. For example, doctors may consider a patient's self-described symptoms, blood test results, x-rays, blood pressure, and stethoscope findings before making a diagnosis. Similarly, ecologists may not be able to classify an organism solely based on visual data, needing some form of audio or genetic data to differentiate similar-looking species. This suggests a more general challenge: how to use different data modalities effectively.

    Multimodal neural networks have been effective in analyzing multiple modalities of data, achieving state-of-the-art performance in a variety of tasks \cite{2021radford, 2025gong}. However, they have two major drawbacks. First, they are fundamentally black-box in nature, which prevents professionals from interpreting the decision-making process of these models. Second, not all data are equal. In practice, data type $A$ (e.g., images of organisms) may be much more accessible than data type $B$ (e.g., genetic sequences), although $B$ may be more informative. Therefore, during inference, we may prefer our models to use only data $A$ if possible and avoid $B$ unless it is absolutely necessary. Doing so can greatly reduce the cost of expensive or invasive procedures with a minimal (or no) decrease in accuracy. Figure \ref{fig:introfig} provides a high level overview of the reasoning process we develop. 

    In computer vision, prototype networks (ProtoPNets) are an emerging approach to designing interpretable neural networks, offering the predictive power of neural networks while providing an inherently interpretable reasoning process \cite{chen2019looks}. ProtoPNet and its variants \cite{nauta2021neural, wang2023learning, donnelly2022deformable, ma2024interpretable, ma2023looks, rymarczyk2021protopshare, rymarczyk2022interpretable, wang2021interpretable, donnelly2025rashomon, waggoner2025creating, wang2025mixture} utilize a case-based reasoning process in which a set of learned prototypical examples of each class is compared to a given input to form predictions.
    We introduce two novel approaches to combine image and genetic modalities, while abstaining from measuring genetic data when possible:

\noindent\textbf{Conformal Abstention Learning (CAL)} CAL is a general technique that can be used beyond image and genetic data. The steps for CAL are as follows: (1) Determine which modality is most expensive. Here, genetic data are most expensive. 
(2) Apply conformal prediction to determine the maximum reasonable change in our predicted logits due to measuring the expensive modality. 
(3) Determine if the maximum reasonable change would be sufficient to alter our overall prediction. If not, we do not need to measure the genetic information.

  \noindent\textbf{Abstention Learning ProtoTree. (ALP)}
   This is a specialized multimodal extension of ProtoTree \cite{nauta2021neural} in which each internal node may consider either a genetic or an image prototype comparison. We leverage ProtoTree's unique tree structure to regularize towards one modality when forming predictions by encouraging entire paths from root to leaf to use only image information.

    We apply CAL and ALP to the BIOSCAN-1M insect classification dataset, which contains paired image and genetic information for 516 insect species.  At inference time, our models consider whether the more expensive modality (genetic information) is necessary to make a prediction; when it is not, they reason solely based on the less expensive modality (in this case, images). 
    We demonstrate that, by identifying cases that are easy to classify with image information alone, 
    we can achieve comparable accuracy to full multimodal models with just a fraction of the genetic information needed overall, all while providing an interpretable reasoning process.

 
\section{Related Works} 

\paragraph{Image-Genetic Species Classification.} Deep neural networks have been successfully deployed to classify species based on image data \cite{martineau2017survey} or genetic data \cite{fluck2022applying, busia2018deep} across a variety of settings. Recently, the research community has expanded on these methods by leveraging both modalities for species classification through a shared embedding space \cite{2025gong}, by ensembling distinct representations of image and genetic data \cite{nanni2025advancing, nanni2025insect}, and by combining both modalities for better out-of-distribution detection \cite{impio2024improving}. However, these methods all follow an opaque reasoning process, raising concerns about their reliability in deployment.

\paragraph{Case-Based Neural Networks.} In an orthogonal line of work, researchers have developed interpretable, case-based deep neural networks for classification tasks \cite{li2018deep, chen2019looks}. Given an input, these models first extract features using a neural network backbone, then make comparisons to a set of learned prototypes using these extracted features. This comparison yields a similarity score for each prototype, and these scores are combined to form a class prediction, yielding an explanation of the form ``this looks like that''. Since the introduction of the first of these models -- called ProtoPNet \cite{chen2019looks} -- a wide variety of improvements and extensions to this framework have been developed \cite{nauta2021neural, wang2023learning, donnelly2022deformable, ma2024interpretable, ma2023looks, rymarczyk2021protopshare, rymarczyk2022interpretable, wang2021interpretable, donnelly2025rashomon, waggoner2025creating, wang2025mixture}. The majority of these models form their classifications using a linear combination of the prototype similarity scores \cite{wang2023learning, donnelly2022deformable, ma2024interpretable, ma2023looks, wang2021interpretable, donnelly2025rashomon, waggoner2025creating, wang2025mixture}. In contrast, ProtoTree \cite{nauta2021neural} learns a decision tree in which each internal node performs a prototype comparison, and each leaf specifies a vector of class logits. 

While these models have primarily been used for image classification, recent extensions have turned to consider additional data modalities. 
Notably, \citet{waggoner2025creating} applied this case based reasoning approach to environmental DNA classification. 
\citet{wang2023novel} extended ProtoPNet to consider both image and text data, while \citet{song2025multimodal} considered text, image, and acoustic modalities. \citet{yang2025protomm} combines four medical data modalities, including images and a table of gene expression levels.
However, no case-based method exists to consider image and raw base-pair genetic sequences (rather than tabular expression levels); moreover, existing multimodal extensions of ProtoPNet assume that every data modality is available for every sample. 
We leverage ideas from conformal prediction and prediction abstention to address this gap.

\paragraph{Abstention and Conformal Predictions} In this work, we draw inspiration from the problem of classification with the option to abstain from making a prediction \cite{chow1959optimum, bartlett2008classification, cortes2016learning}, in which classifiers are encouraged to refrain from making predictions in cases where they are likely to be incorrect. In the same spirit, our aim is to measure genetic information about a sample only when an image classifier abstains from making a prediction. 

We use conformal prediction \cite{lei2014distribution, papadopoulos2002inductive, vovk2005algorithmic} to identify cases in which an image classifier should abstain. Conformal prediction uses a held out calibration set to estimate quantiles for the error of an arbitrary model with respect to a target, providing statistically rigorous prediction intervals. We frame the task of identifying cases where an image classifier should abstain as a multi-output regression task (explained in Section \ref{sec:methods}) and leverage specialized conformal prediction methods for this setting \cite{dheur2025unified}.

\section{Methodology} 
\label{sec:methods}

  \begin{figure*}
    \centering
    \includegraphics[width=\textwidth]{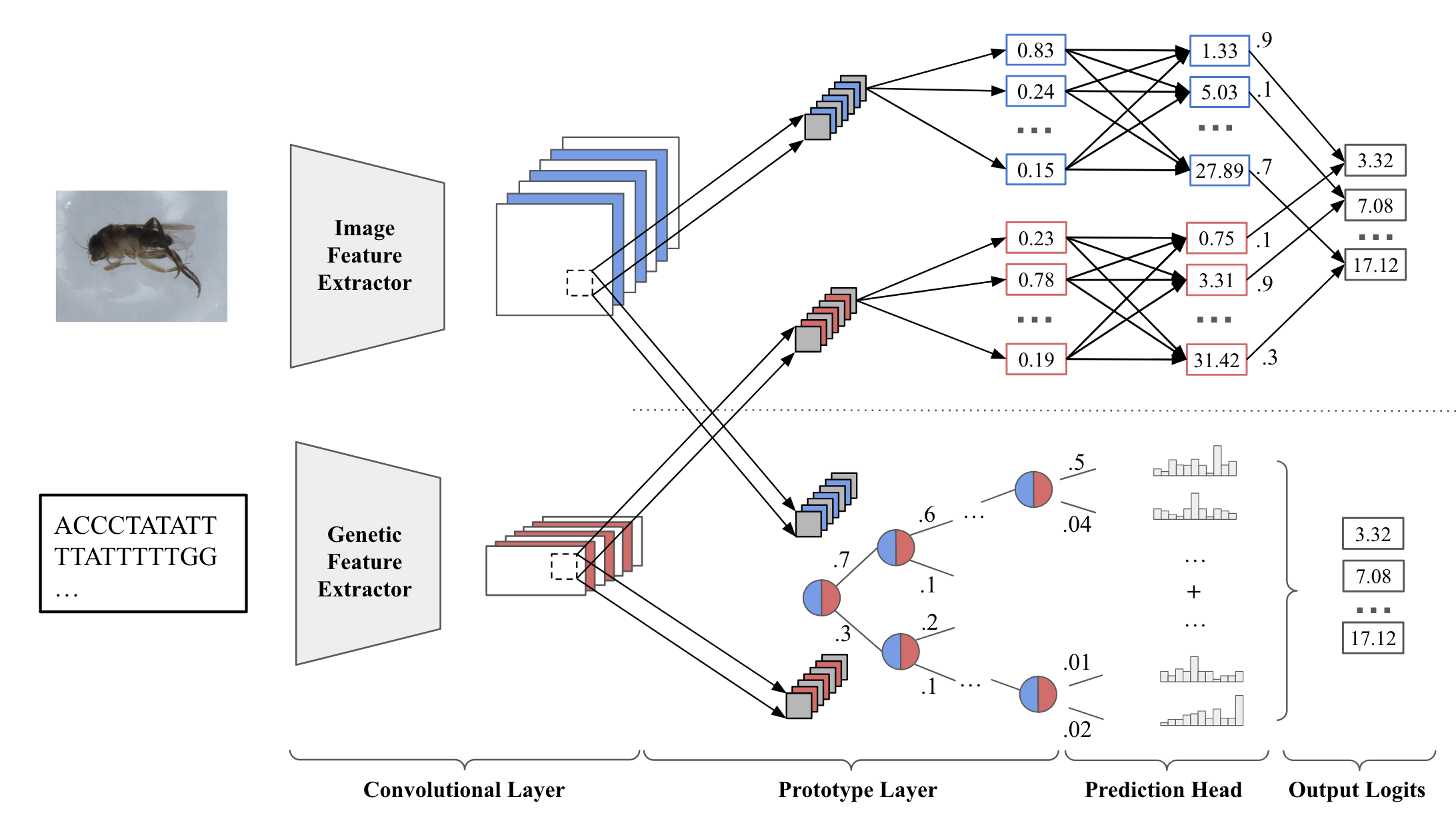}
    \caption{Our multimodal extensions applied to a vanilla ProtoPNet (top) and a ProtoTree (bottom). Both architectures use the same pair of convolutional feature extractors on image and genetic data. In the ProtoPNet, we take a weighted average of the logits using CAL. In the ProtoTree, we consider either a genetic or an image prototype at each node in the tree and traverse the tree based on the prototype's similarity to the input. Some paths from the root to the leaf do not require consideration of any genetic prototypes, while others do. } 
    \label{fig:architecture}
  \end{figure*}


  Let $\mathbf{x} \in \mathbb{R}^{C \times H \times W}$ denote an input image or sequence of height $H$ and width $W$ with $C$ channels, and let $\mathbf{y} \in \{0, \ldots, K-1\}$ denote class labels. For simplicity of notation, we generally define our models and loss terms with respect to a single sample $(\mathbf{x}, \mathbf{y})$ and omit subscripts that distinguish between different samples in a dataset. Following the standard interface of ProtoPNext \cite{2024willard}, a ProtoPNet consists of an embedding layer $f: \mathbb{R}^{C \times H \times W} \to \mathbb{R}^{D \times H^\prime \times W^\prime}$, which extracts a $H^\prime \times W^\prime$ map of $D$ dimensional feature vectors; a prototype layer $g:=g_2 \circ g_1$, where $g_1 :\mathbb{R}^{D \times H^\prime \times W^\prime} \to \mathbb{R}^{P \times H^{\prime\prime} \times W^{\prime\prime}}$ computes the similarity of each of the $P$ learned prototypes to the input at each of $H^{\prime\prime} \times W^{\prime\prime}$ locations and $g_2:\mathbb{R}^{P \times H^{\prime\prime} \times W^{\prime\prime}} \to \mathbb{R}^P$ computes the maximum value over these spatial locations; and a class prediction head $h: \mathbb{R}^{P} \to \mathbb{R}^K$ that uses these prototype similarities to compute an output logit for each class. 
  For conciseness of notation, we define $\mathbf{z} := f(\mathbf{x})$ to be the latent representation of input $\mathbf{x}$, $\mathbf{s} := g \circ f(\mathbf{x})$ to be the maximum similarity between each prototype and the input at any spatial location, and $\hat{\mathbf{y}}:= h \circ g \circ f(\mathbf{x})$ to be a vector of class logits. Broadly, these models minimize a loss function of the form 
  \begin{equation*}
      \ell (f, g, h, \mathbf{x}, \mathbf{y}) = CE(h \circ g \circ f (\mathbf{x}), \mathbf{y}) + \ell^\prime (f, g, h, \mathbf{x}, \mathbf{y})
  \end{equation*}
  where $CE$ denotes the cross-entropy loss, and $\ell^\prime$ is a combination of regularization penalties and interpretability metrics. Furthermore, in almost all ProtoPNet variants, a prototype projection step is performed during and at the end of training to associate each learned prototype with a specific pixel space visualization. 

  We consider the setting where both image and genetic data may be available for a given sample, which we denote by $\mathbf{x}^{(I)} \in \mathbb{R}^{3 \times 224 \times 224}$ and $\mathbf{x}^{(G)} \in \mathbb{R}^{4 \times 1 \times 720}$ respectively. Here, our genetic data is comprised of variable-length strings denoting nucleotides ($A, C, T, G$) that are right-padded, and then one-hot encoded into a tensor in $\mathbb{R}^{4 \times 1 \times 720}$. 
  In each of our approaches, we consider two ProtoPNets, $F^{(I)}:= h^{(I)} \circ g^{(I)} \circ f^{(I)}$ and $F^{(G)}:=h^{(G)} \circ g^{(G)} \circ f^{(G)},$ to separately process image and genetic data, respectively. Figure \ref{fig:architecture} provides an overview of our two approaches in combining the two modalities.


\subsection{Conformal Abstention Learning (CAL)}

  We first consider ensembling $F^{(I)}$ and $F^{(G)}$ through a simple class-wise weighted average of the logits from each model.
%
  For class $j$, the final output logit $\hat{y}_j$ of the multimodal model is a weighted average of the corresponding image and genetic logits:
  \begin{align*}
    \hat{y}_j &= \sigma(m_j) F^{(I)}_j(\mathbf{x}^{(I)}) + (1 - \sigma(m_j)) F^{(G)}_j(\mathbf{x}^{(G)}),
  \end{align*}
  where $m_j \in \mathbb{R}$ is a learnable parameter that determines how much the final output logit depends on its corresponding image and genetic logits. 
  
  We now introduce the application of conformal prediction to this task, then introduce loss terms that aim to make the model weigh one modality more highly than the other.

\subsubsection{Conformal Prediction}
\label{subsec:conformal_prediction}
    \begin{figure}[t]
      \centering
      \includegraphics[width=\columnwidth]{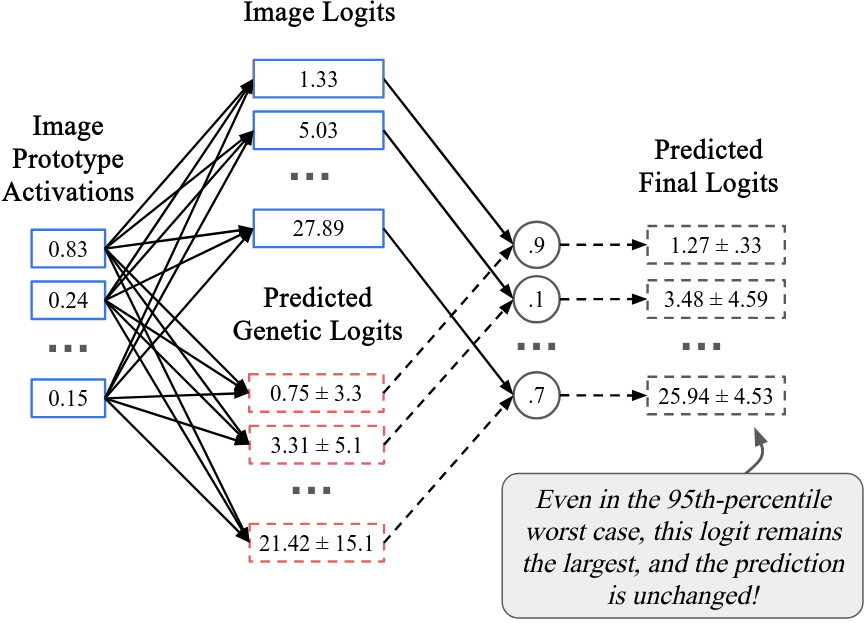}
       \caption{The prediction head of a multimodal ProtoPNet employing conformal prediction to produce prediction sets for CAL. The image prediction head includes an additional fully connected layer to predict the genetic logits, and conformal prediction is employed to create prediction sets around them.}
       \label{fig:cal_architecture}
    \end{figure}

    The construction of our multimodal model separates out the contribution of each genetic logit $F^{(G)}_j(\mathbf{x}^{(G)})$ from that of each image logit $F^{(I)}_j(\mathbf{x}^{(I)})$. As such, if we can bound the values that $(1-\sigma(m_j)) F^{(G)}_j(\mathbf{x}^{(G)})$ could take, we can bound how much the overall value of each logit $\hat{y}_j$ might change relative to $F^{(I)}_j(\mathbf{x}^{(I)})$.

    We leverage split conformal prediction to form these bounds. Given a prediction, split conformal prediction provides a prediction set for its true value with some user-specified confidence value $1-\alpha$. It does so by calibrating the model's residuals on a held-out calibration set to estimate the distribution of prediction errors.
    
    To predict each \textit{genetic} logit, we introduce an additional fully connected layer $\hat h: \mathbb R^{P} \rightarrow \mathbb R^K$ to the \textit{image} model to predict the \textit{genetic} logits $\hat {\mathbf {y}}^{(G)}$. It is initialized with the same weights as $h^{(I)},$ and is trained to minimize the mean squared error between each prediction and $\mathbf {\hat y}^{(G)}$. 
    

     Let $\delta_j$ denote the margin of error of the prediction set for logit $\hat y_j,$
      and let $k$ be the class predicted by $F^{(I)}$ for an input $\mathbf x^{(I)}$.
    Then, with probability at least $1-\alpha$, the genetic data could not yield a more different set of predicted logits than
    \begin{equation*}
    {\tilde y}_j=\begin{cases}
        \sigma(m_j)\hat y_j^{(I)} + (1-\sigma(m_j))(\hat y^{(G)}_j-\delta_j) & \text{if}\ j= k \\
        \sigma(m_j)\hat y_j^{(I)} +(1-\sigma(m_j))(\hat y^{(G)}_j+\delta_j) & \text{otherwise}.
    \end{cases}
    \end{equation*}
    That is, the logit for the predicted class $k$ based on the image data $\hat{y}_k$ is unlikely to decrease past $\tilde{y}_k$, and every other logit $\hat{y}_j$ is unlikely to increase past $\tilde{y}_j$. If $\arg\max \mathbf{\tilde y}=k$, then with $(1-\alpha)$ confidence, we know that introducing genetic information cannot change the prediction from that of the image model. Figure \ref{fig:cal_architecture} visualizes this logic.

    It is worth noting that we compute a distinct marginal conformal interval for each genetic logit; as such, for full statistical rigor, a correction for multiple hypothesis testing should be applied. In practice, we observe that this is not necessary to achieve nominal coverage rates, as demonstrated in Section \ref{sec:experiments}.
    
    Now that we know how to evaluate models for abstention, we turn to how to train models to abstain more often.

\subsubsection{Loss Terms}
The conformal prediction method of Subsection \ref{subsec:conformal_prediction} can be applied to any pair of models, but may not always yield tight intervals. Here, we introduce two loss terms aimed at helping to achieve tighter intervals.
  
To encourage the model to abstain from measuring genetic information when possible, we introduce a 
\textit{margin loss} to increase the distance between the worst-case logits and the decision boundary.
    Let $d_j:=\tilde y_{k}-\tilde y_j$. If $\min_{j \neq k}d_j>0$, then $\tilde y_k$ is still the maximum logit, and the classification is unlikely to be changed by genetic information.
    
    Directly maximizing the margin $\min_{j\neq k}d_j$ passes gradients through a single logit, leading to slow convergence. We instead maximize a soft minimum of $d_j$, defining margin loss as:
\begin{align*}
\ell_\text{margin} 
&:= 
\log\left( \sum_{\substack{j=0, j \ne k}}^{K-1} e^{-d_j} \right).
\end{align*}
Additionally, we can reduce the maximum possible contribution of any given genetic logit by increasing the value of each $\sigma(m_j),$ thereby decreasing the weight $1 - \sigma(m_j)$ assigned to the genetic logit:
\begin{align*}
    \ell_{\text{modality}} := -\sum_{j=0}^{K-1}  \sigma(m_j).
\end{align*}

To train the genetic logit predictor $\hat{h}$, we include a loss term defined as the mean squared error between the true genetic logits and their predicted values.

\begin{align*}
    \ell_\text{predictor}:=\frac{1}{K}\sum_{j=0}^{K-1}\left(\hat y^{(G)}_j-\hat h_j(g^{(I)}(f^{(I)}(\mathbf x_i^{(I)}))\right)^2
\end{align*}

\subsubsection{Optimization} 
To train the multimodal ProtoPNet, we first initialize the model with a ProtoPNet trained for each modality, and freeze all but the modality weights and the fully-connected layers.
We then minimize the overall loss:
\begin{align*}
    \ell_{total} := \ell_{CE} +  \lambda_1 \ell_{\text{modality}}+\lambda_2\ell_{\text{margin}} +\lambda_3\ell_\text{predictor}
\end{align*}
    The cross-entropy loss $\ell_{CE}$ operates on the image logits alone when image data is deemed sufficient at a pre-set confidence value, otherwise it operates on the final multimodal logits.

\subsection{Abstention Learning ProtoTree (ALP)}
  

  Here, we introduce a specialized cost-aware multimodal extension of ProtoTree \cite{nauta2021neural}. We first introduce the original ProtoTree. In ProtoTree, the class prediction head $h^{\text{tree}}$ is structured as a full binary decision tree consisting of a set of $P$ internal nodes $\mathcal{N}$ and a set of leaf nodes $\mathcal{L}$. Each internal node $n \in \mathcal{N}$ is associated with two child nodes, denoted $n_{\text{left}} \in \mathcal{N} \cup \mathcal{L}$ and $n_{\text{right}} \in \mathcal{N} \cup \mathcal{L},$ which are traversed in a soft manner --- that is, the left and right sub-tree of each node are both involved in forming a prediction --- based on $s_n$ for a given input. Each leaf node $l \in \mathcal{L}$ contains an associated logit vector, which we denote $\hat{\mathbf{y}}_l.$ The full prediction head $h^{\text{tree}}$ for a ProtoTree is thus defined recursively as:
  \begin{align*}
      &h^{\text{tree}}(\mathbf{s}):= r(n_{\text{root}}, \mathbf{s}), \text{ where} \\
      &r(n, \mathbf{s}) := \begin{cases}
          (1 - s_n) r(n_{\text{left}}, \mathbf{s}) + s_n r(n_{\text{right}}, \mathbf{s}) & \text{if } n \in \mathcal{N}\\
          \hat{\mathbf{y}}_l & \text{if } n \in \mathcal{L}\\
      \end{cases},
  \end{align*}
  where $n_{\text{root}}$ denotes the root node and $\mathbf{s}$ is the vector of prototype similarity scores. This formula signifies that the output is a weighted average of contributions from the leaves, where each leaf calculation is based on how similar the input is to each prototype at each node along the path to the leaf. In a unimodal ProtoTree, we compute either $h^{\text{tree}}(\mathbf{s}^{(I)})$ or $h^{\text{tree}}(\mathbf{s}^{(G)}),$ depending on the modality. 
  
  As in the original paper, we train using the soft traversal function $h^{\text{tree}}$ but employ hard decision routing at test time by greedily traversing the tree, i.e., go right at internal node $n$ if $s_n > 0.5$ and left otherwise. We call this traversal function $h^{\text{tree,hard}}$.


  In ALP, each node $n$ considers the prototype similarity vector from each modality ($\mathbf{s}^{(I)}$ and $\mathbf{s}^{(G)}$), plus a \textit{modality weight} $m_n \in \mathbb{R}$ that determines how much each modality should contribute to the left/right routing. We define the multimodal similarity (for the multimodal tree) as a combination of the similarities to image and genetic prototypes:
  \begin{equation*}
    s_n^{(I + G)}:= \sigma(m_n) s_n^{(G)}  + (1 - \sigma(m_n)) s_n^{(I)},
  \end{equation*}
  where $\sigma$ is the sigmoid function. Given $\mathbf{s}^{(I + G)}$, the prediction of our multimodal ProtoTree is then computed as $h^{\text{tree}}(\mathbf{s}^{(I+G)}),$ with $h^{\text{tree}}$ defined as before and used in training, and $h^{\text{tree,hard}}$ defined as before, used at test time. 
  

    \begin{figure}
        \centering
        \includegraphics[width=\linewidth]{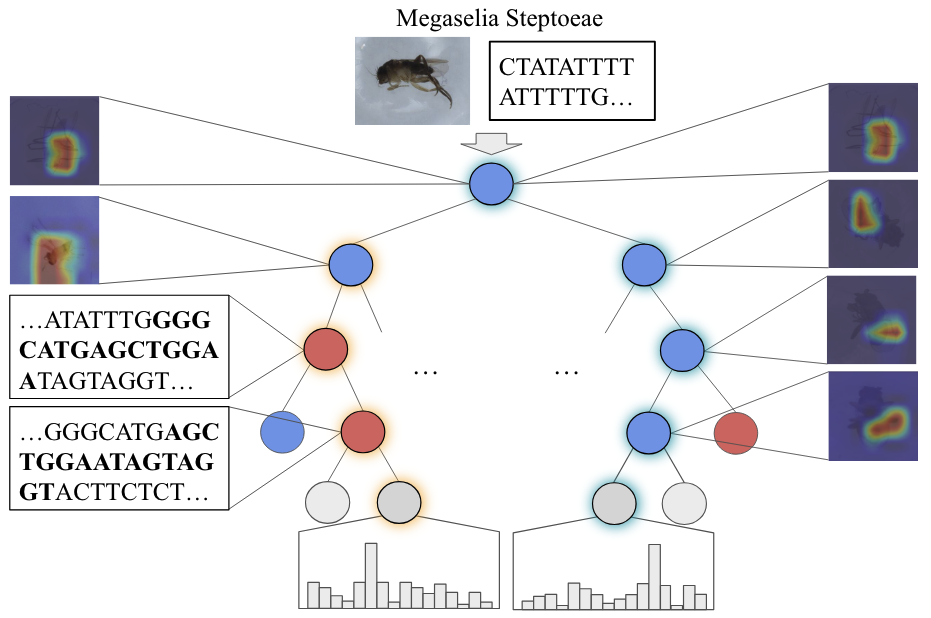}
        \caption{Example reasoning from a trained ALP. Given an input consisting of image and genetic data, we present two possible paths (indicated by highlighted nodes) in which a multimodal ProtoTree may route it to a leaf. In the left path, the first two nodes consider image prototypes and the last two consider genetic prototypes. If the image is routed down this path, we must measure genetic information. The right path, on the other hand, only considers image prototypes. We do not need to measure genetic information for images routed down this path.}
        \label{fig:alp_example}
    \end{figure}

\subsubsection{Threshold Based Modality Assignment}
When one modality is more expensive to collect than the other, users may wish to guide which modality the ProtoTree emphasizes. We leverage a specialized initialization strategy to encourage our ALP model to only use image information when it is sufficient for a strong classification.

  Given a trained image-only ProtoTree $F^{(I)}$ and a leaf node $l \in \mathcal{L}$, let $\mathcal{D}_l$ be the subset of our dataset that is routed to $l$ during a hard traversal of the image-only ProtoTree, and let $\mathcal{N}_l \subset \mathcal{N}$ be the set of all internal nodes in the path from $n_{\text{root}}$ to $l$. For each sample $(\mathbf{x}^{(I)}, \mathbf{y}^{(I)}) \in \mathcal{D}_l$, denote 
  \begin{equation*}
      \textrm{acc}_l = \mathbb{P} \left( F^{(I)} (\mathbf{x}^{(I)}) = \mathbf{y}^{(I)} \mid (\mathbf{x}^{(I)}, \mathbf{y}^{(I)}) \in \mathcal{D}_l \right)
  \end{equation*}
  That is, $\textrm{acc}_l$ denotes the accuracy of a leaf in classifying samples that reach $l$ during a hard traversal.
  We aim to keep all image nodes that lead to accurate classification and replace the ineffective ones with genetic nodes by defining a threshold $t \in [0, 1]$, and initializing the set of internal nodes
  \begin{equation*}
      \mathcal{N}_t = \{n \in \mathcal{N} \mid n \in \mathcal{N}_l, \text{acc}_l > t\}
  \end{equation*}
  that contribute to these high-accuracy image decisions to have image-only modalities. That is, we set $m_n = \tau$ if $n \in \mathcal{N}_t,$ and $m_n = \tau$ otherwise 
  for some positive value $\tau$.
  By doing so, we preserve the performance of the unimodal image ProtoTree for $n \in \mathcal{N}_t$, whilst maximally exploiting the genetic prototypes for all subsequent splits. 
  A natural value for the threshold $t$ is the accuracy of the genetic unimodal ProtoTree, which indicates that we only want to keep nodes that lead to predictions that are at least as confident as the genetic ProtoTree. If we set $\tau$ to a very large value, then it is equivalent to deleting the subtrees whose leaves all have low accuracy; those subtrees will be replaced with genetic subtrees.

  

\subsubsection{Deterministic Modality Routing}

  During training of the multimodal tree, which is trained from both image and genetic data, we do not constrain $m_n$, but we apply a regularization term encouraging the modality weights of each node to polarize towards $+\infty$ or $-\infty$ so that only one modality will be used. We apply an entropy-based polarization term to the probability space as: 
  \begin{align*}
    \ell_{routing} = \frac{1}{P}\sum_{n=0}^{P-1}\bigg( &\sigma(m_n) \log(\sigma(m_n)) \\
    &+ (1 - \sigma(m_n)) \log{(1 - \sigma(m_n))\bigg)}.
  \end{align*} 
  To maintain simple interpretations, we  clip the modality weights $m_n$ to $\pm\infty$ at the end of training to ensure that at each split, an internal node will either solely observe an image prototype or a genetic prototype. 
  
\subsubsection{Optimization Objective} 
To train the multimodal ProtoTree, we first trained two models: an image-only ProtoTree and a genetic ProtoTree. 
We initialize the multimodal ProtoTree using the convolutional weights and prototypes from both networks. We also use the image-only ProtoTree to initialize our modality weights using threshold based modality assignment. 
For paths that end with an image prototype, we initialize the leaf in that path using the corresponding logit vector from the image ProtoTree; otherwise, we use the logit from the genetic ProtoTree.
We then freeze the image-only ProtoTree weights and train the genetic part of the multimodal  model to minimize the overall loss: 
    \begin{align*}
    \begin{split}
        \ell_{total} = & \ell_{CE} + \lambda_1 \ell_{cluster} + \lambda_2 \ell_{ortho} + \lambda_3 \ell_{var} \\
        &+ \lambda_{4} \ell_{wd} + \lambda_{5} \ell_{routing}
    \end{split}
    \end{align*}
    where $\ell_{cluster}$ is the cluster loss from \cite{nauta2021neural}, $\ell_{ortho}$ is the orthogonality loss from \cite{wang2021interpretable}, $\ell_{wd}$ is weight decay, and  $\ell_{var}$ is a variability loss that aims to encourage diversity among prototype activations, defined as
    \begin{align*}
        \ell_{var} = -\frac{1}{H^{\prime\prime}W^{\prime\prime}}\sum_{h=1}^{H^{\prime\prime}}\sum_{w=1}^{W^{\prime\prime}} \mathbb{V}[(g_1 \circ f(\mathbf{x}))_{h, w}],
    \end{align*} 
    where $\mathbb{V}$ denotes the finite sample variance taken over the $P$ entries in $(g_1 \circ f(\mathbf{x}))_{h, w}$. 
    We perform prototype projection at the end of training, as defined in \cite{chen2019looks}. 
 
\section{Experiments}
\label{sec:experiments}
    We evaluated the performance of our multimodal models for species classification using the BIOSCAN-1M insect classification dataset \cite{2023gharaee}. Though the dataset contains more than 1 million samples, only about 84,000 are labeled with a species. We employed a 60/20/20 train/validation/test split. For training our CAL model, the validation set was also used as the calibration set for our conformal intervals. The species distribution is extremely unbalanced and heavy-tailed, so we proportionally oversampled minority classes during training, and evaluated our results primarily on balanced accuracy. To evaluate our ability to prioritize using only image data, we introduce a new metric: success rate. Success rate is defined as the percentage of samples classified using exclusively image data, where a higher success rate indicates more samples were classified without genetic data. See the supplement for more details regarding our experimental setup. 

    We used a ResNet-50 backbone pretrained on iNaturalist for our image modality, and a shallow CNN (described in the supplement) backbone for our genetic modality.
    Our CAL and ALP models were initialized with trained image and genetic ProtoPNet and ProtoTree models, respectively.
    A more detailed set of hyperparameter and architecture settings is provided in the supplement.

    We optimized each ProtoPNet, including baseline models, using Bayesian hyperparameter optimization for 36 GPU hours, allowing runs that started before the time limit to train to completion. We evaluate a multimodal ProtoPNet as an instance of CAL, and a multimodal ProtoTree as an instance of ALP. 

    
    \begin{table}
        \centering
        \begin{tabular}{|c|c|c|}
            \hline
            & \textbf{Image} & \textbf{Genetic}  \\ 
            \hline
            \textbf{ResNet-50} & 78.32 (100\%)& - \\
            \textbf{Genetic CNN} & - & 98.64 (0\%) \\
            \textbf{BioCLIP} \cite{stevens2024bioclipvisionfoundationmodel} & 50.39 (100\%) & - \\
            \textbf{CLIBD} \cite{2025gong} & 51.39 (100\%) & 95.79 (0\%)\\
            \hline 
            \textbf{ProtoPNet \cite{chen2019looks}} & 79.06 (100\%) & 98.16 (0\%)  \\ 
            \textbf{ProtoTree \cite{nauta2021neural}} & 67.59 (100\%) & 87.50 (0\%)  \\ 
            \textbf{Deformable \cite{donnelly2022deformable}} & 78.71 (100\%) & \textbf{99.24} (0\%) \\ 
            \textbf{Support Trivial \cite{wang2023learning}} & 57.93 (100\%) & 88.57 (0\%)  \\ 
            \hline
            \textbf{CAL} ($\alpha$=0.00) & \multicolumn{2}{c|}{{97.57} (0\%)} \\ 
            \textbf{CAL} ($\alpha$=0.05) & \multicolumn{2}{c|}{97.32 (40.26\%)} \\ 
            \textbf{ALP} (t=0.8) & \multicolumn{2}{c|}{82.60 (76.15\%)} \\ 
            \textbf{ALP} (t=0.6) & \multicolumn{2}{c|}{78.14 (84.73\%)} \\ 
        \hline
        \end{tabular}
        \caption{Accuracy comparison to prior work for species classification on BIOSCAN-1M Dataset. We report the balanced accuracy of each model on the test split, which contains 8,200 samples. The success rate of each model is reported in parentheses. The highest accuracy is in bold. While a genetic Deformable ProtoPNet achieves the highest accuracy, CAL achieves only 1.9\% less accuracy while finding a success rate of 40.26\%. 
        }
        \label{tab:benchmarks}
    \end{table}


    \begin{figure*}[h!]
        \centering 
        \includegraphics[width=0.8\linewidth]{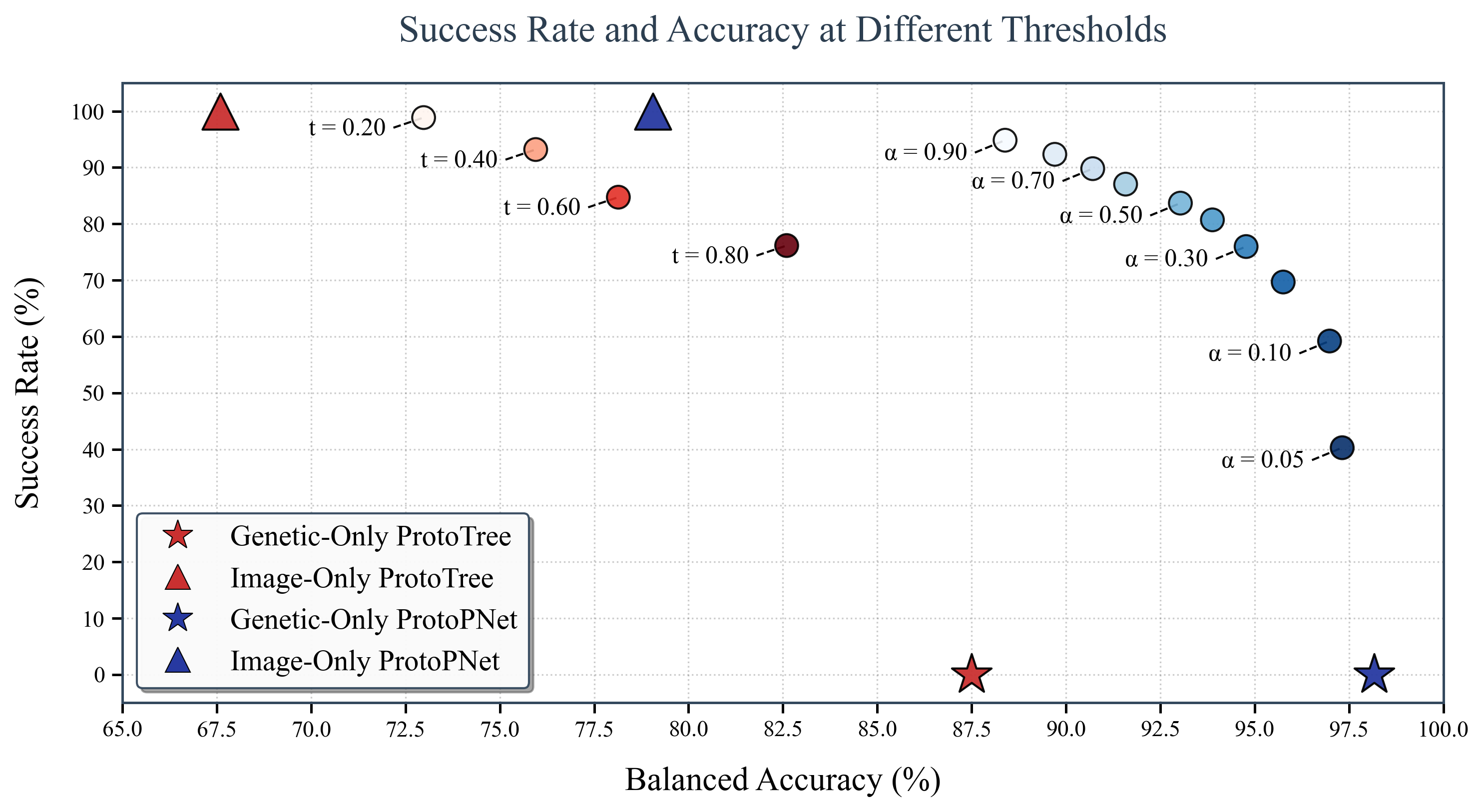}
        \caption{Balanced accuracy vs success rate for ALP and CAL models. \textcolor{red}{Red} is for ProtoTree and \textcolor{blue}{blue} is for ProtoPNet. 
        We see that there is a very small loss in accuracy for a very large boost in success rate at small values of $\alpha$ for CAL; with $\alpha=0.1,$ there is a 0.5\% decrease in accuracy from the most accurate ProtoPNet, in exchange for 59.20\% success rate. In ALP, we see that there is a larger trade-off, with a 5\% decrease in accuracy for 76.15\% success rate, but that ALP is effective in obtaining high success rates.} 
        \label{fig:activations}
    \end{figure*}

    \subsection{Quantitiative Results}
    \textbf{Our experiments show that CAL and ALP can achieve high balanced accuracy while querying less genetic data.}
    As shown in Figure \ref{fig:activations}, both our multimodal ProtoPNet and ProtoTree can achieve competitive accuracies with their genetic counterparts while using substantially less genetic data. For the multimodal ProtoPNet at a confidence value $\alpha = .05$, we see a drop in balanced accuracy of only 0.84\% compared to a genetic ProtoPNet, while gaining a 40.26\% success rate, meaning we are able to classify 40.26\% of the samples using only image information. In other words, this means that in 40.26\% of cases, we would \textit{not} have to collect expensive genetic data to make a classification. 
    We see a moderate loss in accuracy (4.9\%) between multimodal ProtoTree and genetic ProtoTree in exchange for a substantial 76.15\% improvement in success rate with ALP.


    \begin{table}
        \centering
        \begin{tabular}{|c|c|c|}
            \hline
            & \textbf{Balanced Accuracy} & \textbf{Success \%}\\
            \hline
            \textbf{Mar. + Mod. Loss} & \textbf{97.32} & 40.26\% \\
            \textbf{Mar. Loss} & 96.65 & 37.65\% \\
            \textbf{Mod. Loss} & 94.14 & \textbf{61.61}\% \\
            \textbf{Neither Loss} & 97.13 & 10.39\% \\
            \hline
        \end{tabular}
        \caption{Ablation on CAL for margin and modality loss. Metrics are reported for the best-performing CAL models with and without each loss. Metrics are evaluated at a confidence value of $\alpha=0.05$. Mar. is margin loss, and Mod. is modality loss. Including both loss terms yields the best accuracy, with the second best success rate.}
        \label{tab:cal_ablation}
    \end{table}

    \begin{table}
        \centering
        \begin{tabular}{|c|c|c|}
            \hline
            & \textbf{Balanced Accuracy} & \textbf{Success \%}\\
            \hline
            \textbf{Var. + Rout. Loss} & \textbf{82.60} & \textbf{76.15\%} \\
            \textbf{Var. Loss} & 81.75 & \textbf{76.15\%} \\
            \textbf{Rout. Loss} & 78.17 & 74.39\% \\
            \textbf{Neither Loss} & 77.84 & 74.14\% \\
            \hline
        \end{tabular}
        \caption{Ablation on ALP for variability and routing loss.  Metrics are reported for the best-performing ALP model with threshold $t = 0.8$, with and without each loss. Var is variability loss, and Rout is routing loss. Including both loss terms yields the best accuracy and the best success rate.}
        \label{tab:alp_ablation}
    \end{table}
    
    \textbf{Our models produce comparable accuracy to baseline genetic only models.}
    As seen in Table \ref{tab:benchmarks}, both our multimodal ProtoPNet and ProtoTree models perform similarly to existing genetic models and consistently outperform existing image-based species classification models. We see that CAL can achieve 97.57\% balanced accuracy, competitive with genetic-only black-box models like a Genetic CNN at 98.64\% and other interpretable models like vanilla ProtoPNet at 98.16\%, while being slightly worse than Deformable ProPNet at 99.24\%. For ALP, we see that it only gets a balanced accuracy of 82.60\%; this mirrors the relatively low balanced accuracy of both image and genetic ProtoTree compared to other models. While we note that ALP does perform worse than the pure genetic models, this comes with a massive boost in success rate.

    \textbf{Our ablation experiments indicate that our optimization terms improve either accuracy or success rates.} 
    We performed an ablation study on the different losses to both the CAL and ALP models. First, we ablate the margin and modality loss for our CAL model in Table \ref{tab:cal_ablation}. Here, we see that with the addition of either margin or modality loss, we lose balanced accuracy, but increase success rate. However, by   introducing \textit{both} the margin loss and modality loss,  we see the highest accuracy across settings, and the second highest success rate, validating the use of both terms together. 
    In Table \ref{tab:alp_ablation},  we see that the ALP model improves in both balanced accuracy and success rate with the addition of each loss term individually. This shows that all of our loss terms contribute to increasing the accuracy of our model without trading off our ability to identify which samples we can classify using image data alone. 
    
    \begin{figure}
        \centering
        \includegraphics[width=0.8\linewidth]{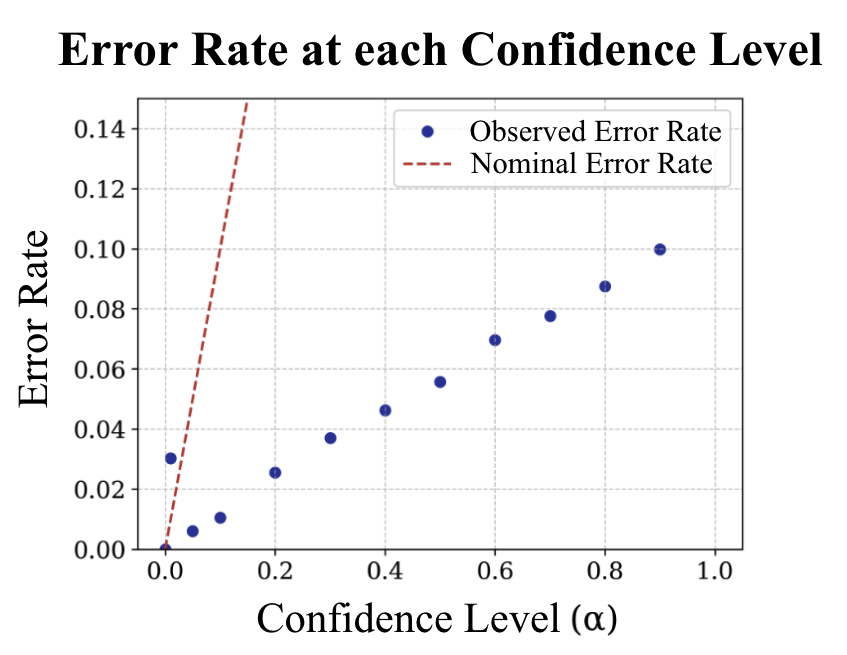}
        \caption{Error rate at each confidence level for CAL models. Here, we see that the error rate is often far below the target confidence level, meaning CAL rarely incorrectly abstains from gathering genetic information. This kind of error happens far below the nominal rate.}
        \label{fig:placeholder}
    \end{figure}

    \textbf{The confidence from our conformal intervals is well calibrated.} 
    Figure \ref{fig:placeholder} shows that, for a wide range of values of $\alpha$, the proportion of samples on which CAL incorrectly says we can abstain from genetic information falls substantially below the nominal rate.
    We see that the error rate scales roughly linearly with our $\alpha$ level, going only up to a 10\% error rate with an $\alpha=.9$.  
    We observe only a single outlier at $\alpha=0.01$, where the error rate slightly exceeds the nominal rate.
    These results show that our intervals achieve nominal coverage, even with marginal conformal intervals, with even high $\alpha$ values showing a relatively small error rate. This implies that no adjustment for simultaneous coverage of logits is necessary in practice; nonetheless, in the supplement,  we show results using a more rigorous statistical determination of confidence for our conformal intervals.
    

    \subsection{Qualitative Results}
        Here, we analyze an example reasoning process from each of CAL and ALP.
        Figure \ref{fig:cal_reasoning_1} presents an example of the reasoning process of CAL with $\alpha=0.05$. Given an input image, CAL first compares the image to a set of image prototypes and applies conformal prediction to decide whether genetic information is necessary. When classifying the Xylosandrus crassiusculus in Figure \ref{fig:cal_reasoning_1} (Top), CAL identifies that genetic information could possibly sway its prediction, and measures the specimen's genetic sequence. When classifying the Argyria centrifugens in Figure \ref{fig:cal_reasoning_1} (Bottom), the margin between the predicted class logit and the next highest logit is sufficiently high that, with high probability, the genetic information would not change the models prediction from Argyria centrifugens.
        
        Figure \ref{fig:alp_example} provides an example reasoning process for ALP. Given an input image --- in this case, a Megaselia steptoeae --- ALP computes the similarity of the prototype at the root node to the image; if the similarity is high, it traverses right, otherwise it traverses left. This process repeats until either (1) in the case of the right path in Figure \ref{fig:alp_example}, a leaf node is hit, yielding a prediction without genetic information, or (2) in the case of the left path, a genetic prototype is hit, confirming that we must measure genetic information. In this case, ALP traverses the right path and correctly classifies the specimen as a Megaselia steptoeae using only image information.


        
        
        
        

    


\section{Conclusion} 

    In this work, we presented two multimodal extensions of ProtoPNet: Conformal Abstention Learning (CAL), and Abstention Learning ProtoTree (ALP). We showed that these methods match the strong performance of multimodal black box models while providing interpretability and often avoiding the need to measure expensive genetic data. Moreover, our conformal prediction framework introduced for Multimodal ProtoPNet is modular and can be applied to ensemble any unimodal models in such a way that expensive modalities can often be avoided. 
    
    The ability to avoid expensive data modalities has substantial positive impacts in practice. When monitoring rare or endangered species, it may allow researchers to avoid capturing or harming individuals of the target species, thereby preventing the resulting ecological harm. For medical applications, doctors may be able to avoid running expensive or invasive tests.
   
    
    While the conformal prediction approach used in CAL is effective, it is often overly conservative because we used a lightweight conformal prediction approach. Future work should consider more advanced conformal prediction methods. Additionally, this work considered two data modalities; in practice, there may be several modalities of interest, each with varying costs. Future studies should consider extending this abstention-based approach to larger numbers of modalities with varying associated costs. 


%
{
    \small
    \bibliographystyle{ieeenat_fullname}
    \bibliography{main}
}

\clearpage
\section*{Appendix}

\subsection{Additional Examples of Reasoning Processes}


\subsubsection{Conformal Abstention Learning}
\begin{figure*}
        \centering
        \includegraphics[width=\linewidth]{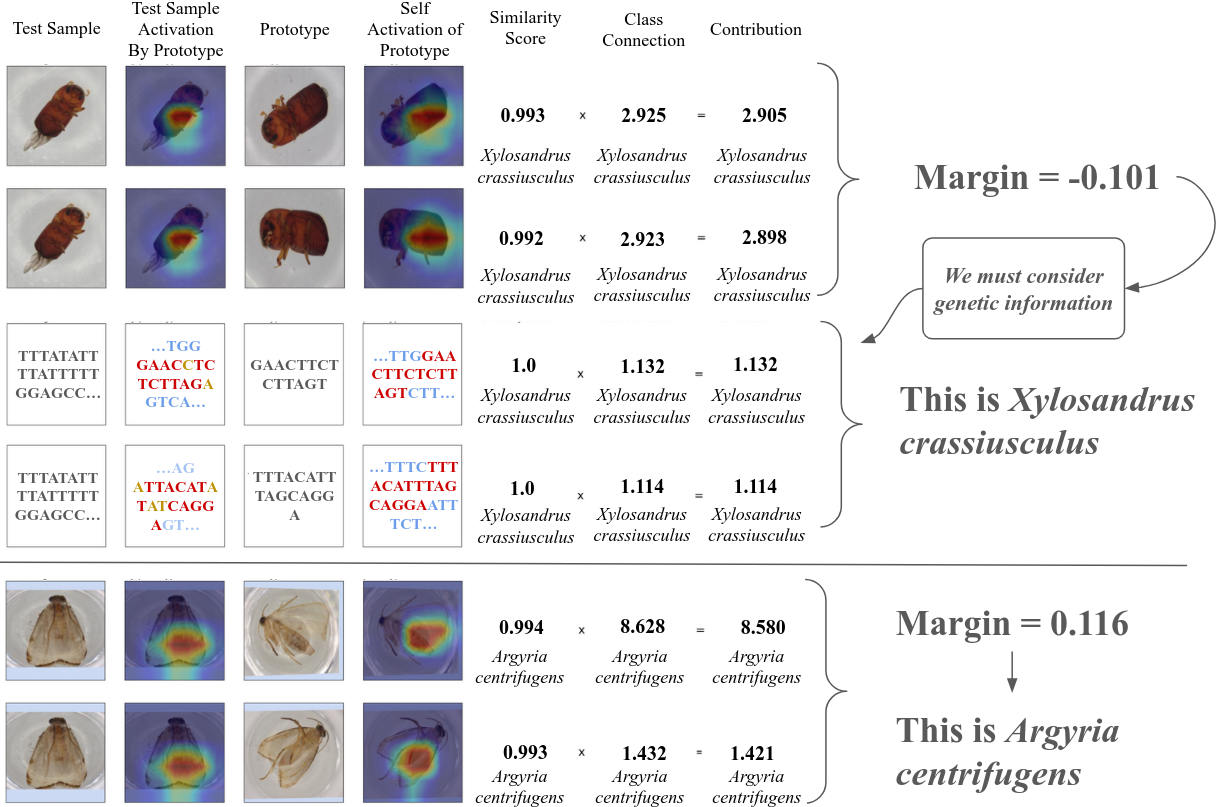}
        \caption{Example of a CAL reasoning process with $\alpha=.05$. For the top sample, both image and genetic information are required. For the bottom sample, only image information is required.}
        \label{fig:cal_reasoning_1}
    \end{figure*}

\begin{figure*}
        \centering
        \includegraphics[width=\linewidth]{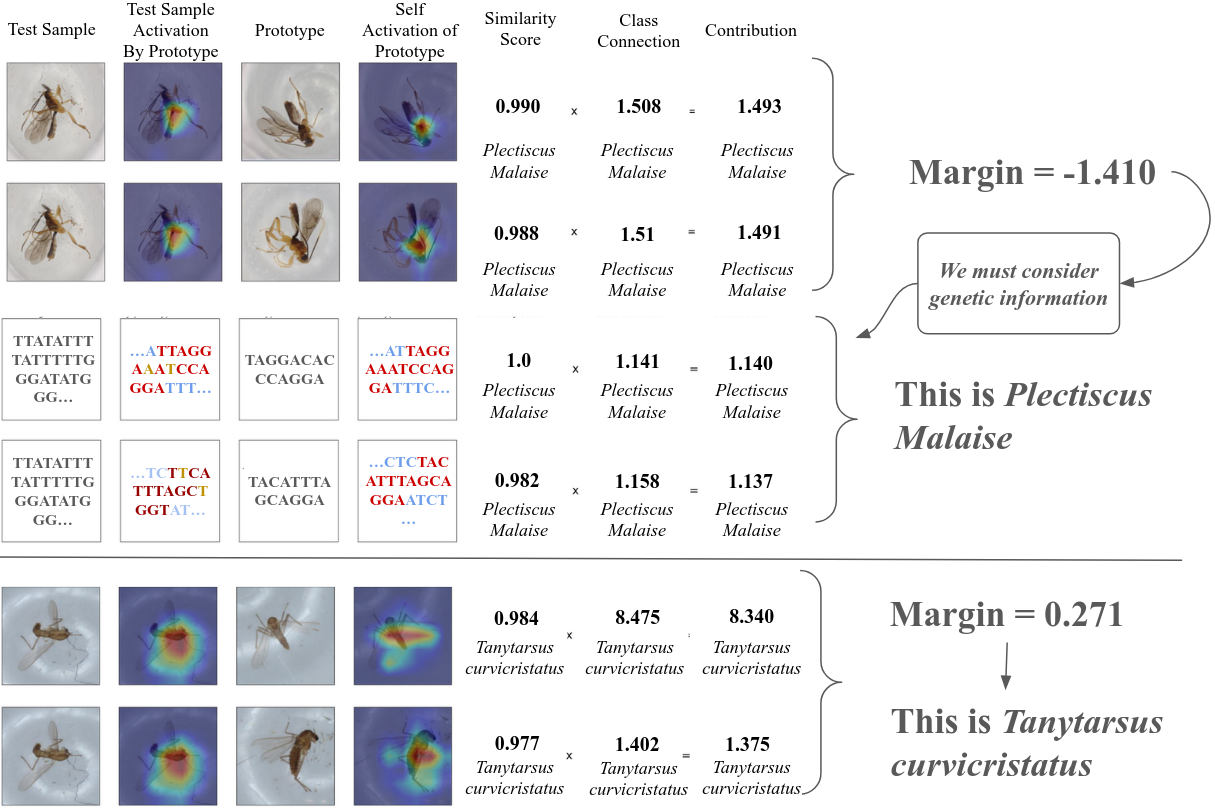}
        \caption{Another example of a CAL reasoning process with $\alpha=.05$. For the top sample, both image and genetic information are required. For the bottom sample, only image information is required.}
        \label{fig:cal_reasoning_2}
    \end{figure*}

    In Figure \ref{fig:cal_reasoning_1}, we present two examples of the reasoning process of the Conformal Abstention Learning (CAL) ProtoPNet. The first involves querying genetic information while the second does not. In Figure \ref{fig:cal_reasoning_2}, we show another pair of examples, one of which involves genetic data.

    In all four examples, the image network alone correctly classifies the sample. But, for two of the samples, CAL identifies that genetic information could change the prediction.

\subsubsection{Abstention Learning ProtoTree}

\begin{figure*}
        \centering
        \includegraphics[width=\linewidth]{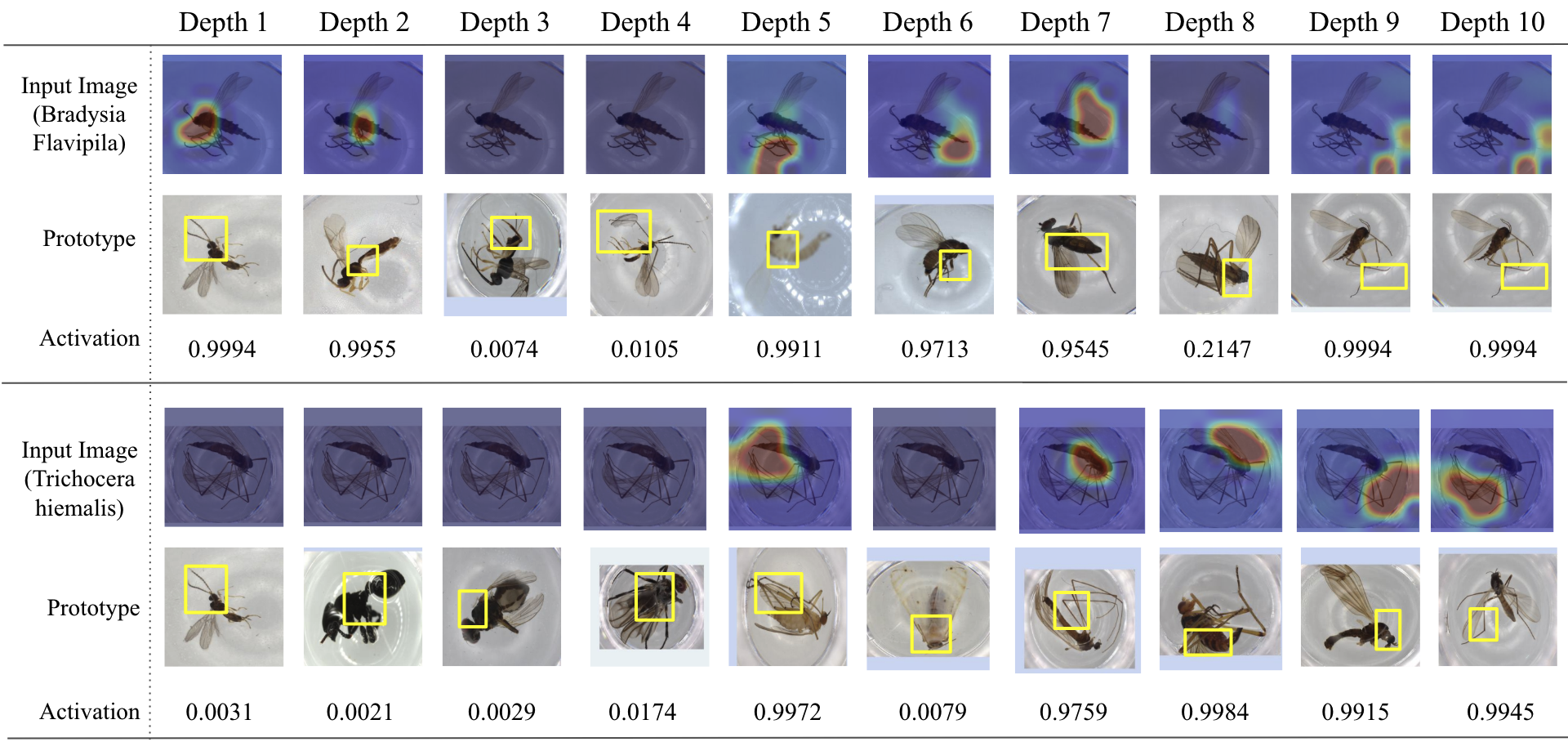}
        \caption{Reasoning path of ProtoTree for samples that require only image prototypes. Both input samples are correctly classified by the model.}
        \label{fig:reasoning_prototree_imgonly}
    \end{figure*}

\begin{figure*}
        \centering
        \includegraphics[width=\linewidth]{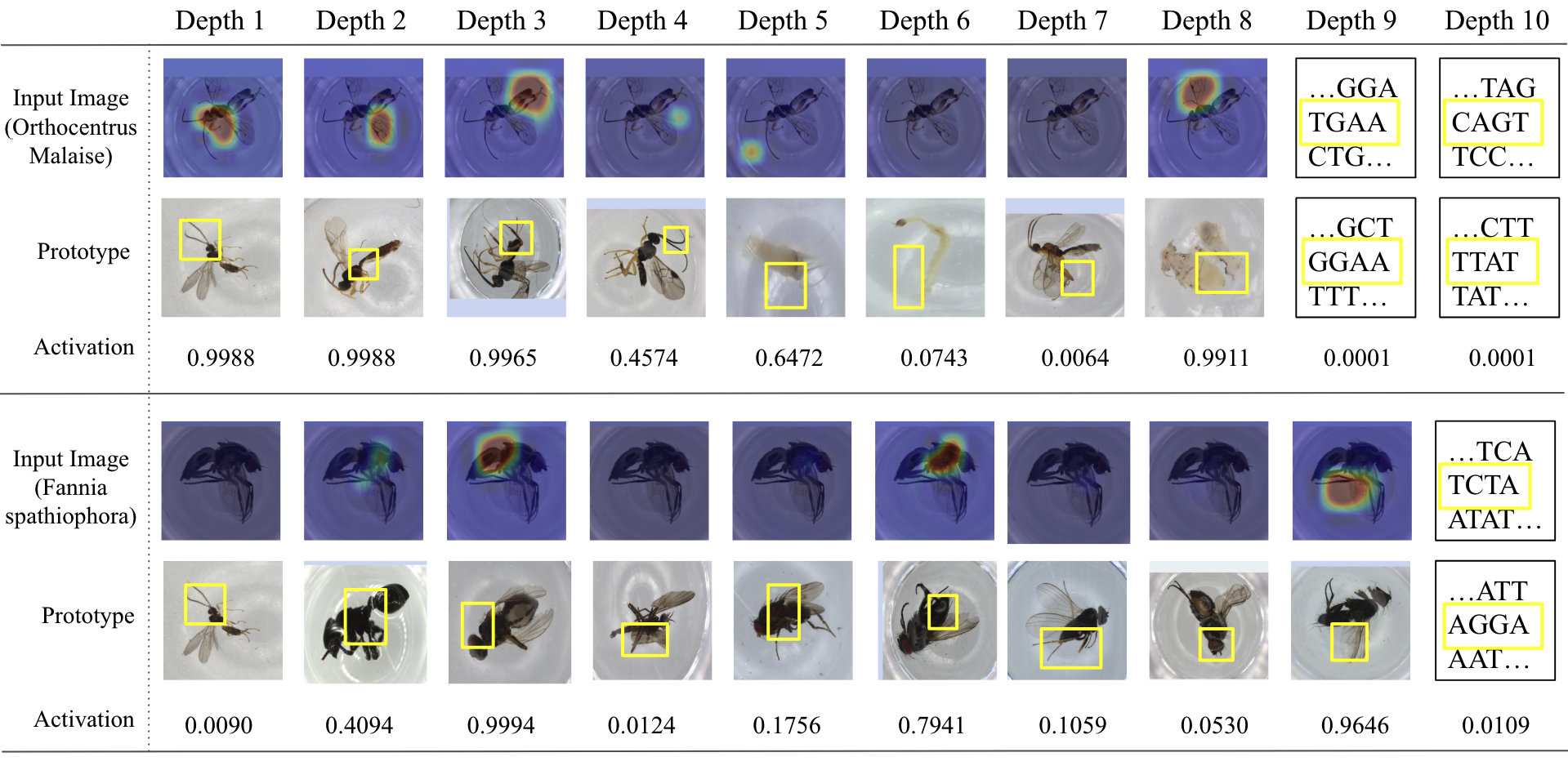}
        \caption{Reasoning path of ProtoTree for samples that require genetic prototypes. Both input samples are correctly classified by the model.}
        \label{fig:reasoning_prototree_genetic}
    \end{figure*}

    We present additional example reasoning processes for the Abstention Learning ProtoTree (ALP). In Figure \ref{fig:reasoning_prototree_imgonly}, we show two correctly-classified samples that require querying only image data, and in Figure \ref{fig:reasoning_prototree_genetic}, we show two correctly-classified samples that require querying at least one instance of genetic data. During the hard traversal, each internal node---which has either an image prototype or genetic prototype---computes the maximum activation of its image/genetic prototype with the latent patches generated by the convolutional feature extractor. If the activation is greater than $0.5$, the sample gets routed to the right child node, and left otherwise. 

    The majority of the activations in both images and genetic sequences are polarized, as in they are close to $0$ or $1$. This indicates that the model, in most cases, is confident about the presence (or lack thereof) of a prototype in an image. The concentrated portion of red on each heatmap reveals the location where a prototype activates highly on an input image. 

\section{Image Backbone Benchmarks}

    For the image ProtoPNets, we ran additional benchmarks on DenseNet and VGG-16. We report the results in Table \ref{tab:backbone_acc}. 

    \begin{table}[h]
        \centering
        \begin{tabular}{l c c}
        \hline
        \textbf{Model} & \textbf{Blackbox} & \textbf{ProtoPNet} \\
        \hline
        ResNet-50     & 78.32 & 79.06\\
        Densenet-161     & 84.01 & 80.47\\
        VGG-16          & 70.05 & 72.36\\
        \hline
        \end{tabular}
        \caption{Balanced accuracies (\%) of blackbox and ProtoPNet modles using different backbone architectures. }
        \label{tab:backbone_acc}
    \end{table}

    We further trained CAL ProtoPNet models using these ProtoPNet models for initialization. The success rate and accuracy of these models are reported in Figure \ref{tab:conformal_backbone_acc}

    \begin{table}[h]
        \centering
        \begin{tabular}{l c c}
        \hline
        \textbf{Backbone} & \textbf{Balanced Accuracy} & \textbf{Success \%} \\
        \hline
        ResNet-50     & 97.32 & 40.26\\
        Densenet-161     & 95.26 & 39.17\\
        VGG-16          & 95.22 & 26.62\\
        \hline
        \end{tabular}
        \caption{CAL ProtoPNet balanced accuracies (\%) and success rates (\%) using different backbone architectures at $\alpha=.05$.}
        \label{tab:conformal_backbone_acc}
    \end{table}

\section{Multi Conformal Prediction}

    For full statistical rigor, we introduce a variation of CAL which uses M-CP to estimate a confidence region around the predicted genetic logit vectors \cite{dheur2025unified}. To do this we apply CAL using the $L_\infty$ distance between the true and predicted genetic logits as a conformity score.
    
    \begin{figure}
        \centering
        \includegraphics[width=0.9\linewidth]{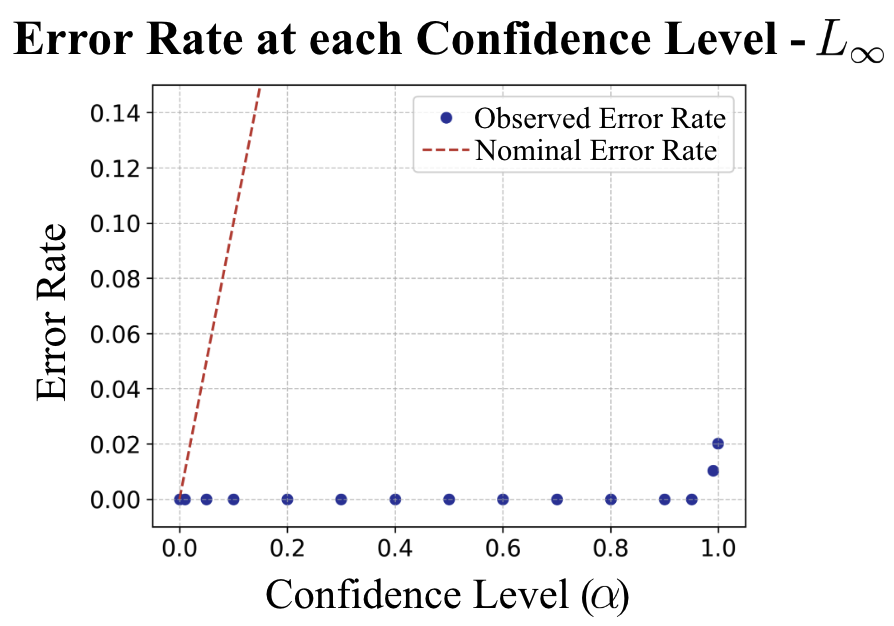}
        \caption{Error rate at each confidence level for the $L_\infty$ CAL model. Here, we see that M-CP CAL rarely incorrectly abstains from gathering genetic information.}
        \label{fig:linf-error}
    \end{figure}
    
    \begin{figure}
        \centering
        \includegraphics[width=0.9\linewidth]{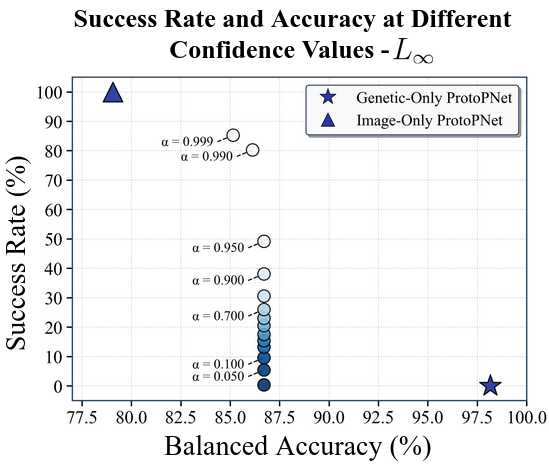}
        \caption{Success rate vs. balanced accuracy for our M-CP CAL model. The model performs approximately 12\% worse than a plain genetic ProtoPNet model at most confidence values.}
        \label{fig:linf-acc}
    \end{figure}
    
    Although rigorous, this approach results in overly conservative confidence regions. In order to classify without genetic data, the model must be heavily regularized to favor image information. The accuracy suffers as a consequence, as shown in Figure \ref{fig:linf-acc}.
    
    Since the bounds are so conservative, increasing the confidence value $\alpha$ has no effect on accuracy up to $\alpha=0.95$. That is, up to $\alpha=0.95$, the model never mistakenly forgoes genetic information, as shown in Figure \ref{fig:linf-error}.

\section{Experimental Setup}


\subsection{Genetic Backbone Architecture}

    The specifications of the genetic backbone are detailed in Table \ref{tab:genetic_backbone}. We noticed that a deeper architecture was not necessary to achieve high balanced accuracy, so we opted for this relatively shallow network. 

    \begin{table*}[h]
        \centering
        \begin{tabular}{lcccccc}
        \hline
        \textbf{Layer} & \textbf{Type} & \textbf{Input} & \textbf{Output} & \textbf{Kernel} & \textbf{Stride} & \textbf{Padding} \\
         & & \textbf{Channels} & \textbf{Channels} & \textbf{Size} & & \\
        \hline
        Conv1 & Conv2D & 4 & 16 & $(1, 3)$ & $(1, 1)$ & $(0, 1)$ \\
         & ReLU & 16 & 16 & -- & -- & -- \\
        Pool1 & MaxPool2D & 16 & 16 & $(1, 3)$ & $(1, 3)$ & $(0, 0)$ \\
        \hline
        Conv2 & Conv2D & 16 & 32 & $(1, 3)$ & $(1, 1)$ & $(0, 1)$ \\
         & ReLU & 32 & 32 & -- & -- & -- \\
        Pool2 & MaxPool2D & 32 & 32 & $(1, 3)$ & $(1, 3)$ & $(0, 0)$ \\
        \hline
        Conv3 & Conv2D & 32 & 64 & $(1, 3)$ & $(1, 1)$ & $(0, 1)$ \\
        Pool3 & MaxPool2D & 64 & 64 & $(1, 2)$ & $(1, 2)$ & $(0, 0)$ \\
        \hline
        \end{tabular}
        \caption{Architecture of convolutional backbone for genetic data. }
        \label{tab:genetic_backbone}
    \end{table*}

\subsection{Dataset Processing}

    In the BIOSCAN-1M dataset, there are a total of 84,000 samples labeled at the species level. We use a train/validation/test split of 60/20/20, with a minimum of 10 samples per class, which leaves us with a dataset of 41,660 samples over 516 classes. Due to the heavy class imbalance of this dataset, we evaluate our models using balanced accuracy. During training, underrepresented classes are over-sampled so that the model sees an equal distribution of samples from all classes.
    
    We perform image augmentation during training by introducing random rotations, skews, shears, flips, and distortions. All images are normalized and cropped to the shape $(3, 224, 224)$. The variable-length raw genetic data are sequentially one-hot encoded to $C=4$ channels (A, C, T, G) with unknown or missing base pairs encoded into $0 \in \mathbb{R}^4$. We then augment each genetic sequence by introducing random genetic substitutions, insertions, and deletions---analogous to point and frameshift mutations in biology. Finally, the sequence is  zero-padded to preserve consistent width, resulting in a shape of $(4, 1, 720)$.
    Since the absolute position of nucleotides within a genetic sequence is meaningful, we add a sinusoidal positional encoding to the latent space.

\subsection{Hyperparameters}

\subsubsection{Conformal Abstention Learning} 

    When training a CAL ProtoPNet model, we initialize the constituent models for each modality with pretrained ProtoPNets. We initialize $\hat h$, the predictor linear layer, with the weights of the image linear layer $h^{(I)}$. We then train only the linear layers and modality weights.

    We optimize hyperameters using a Bayesian optimization scheme. The descriptions and prior distributions of all hyperparameter values can be found in Table \ref{tab:sweep_multimodal}. 
    
    We seek to maximize the following objective in our optimization on the validation set:
    \begin{equation*}
        \left(\text{Balanced Accuracy}_{\alpha=0.05} - 0.9\right)\times \text{Success Rate}_{\alpha=0.05}
    \end{equation*}

    This objective was chosen to ensure that success rate and accuracy are simultaneously high and that the overall balanced accuracy of the model does not decrease far past that of the genetic ProtoPNet.
        
    \begin{table*}[h!]
        \centering
        \footnotesize
        \begin{tabular}{l l l l p{5cm}}
        \hline
        \textbf{Hyperparameter Name} & \textbf{Distribution} & \textbf{ProtoPNet (Genetic)} & \textbf{ProtoPNet (Image)} & \textbf{Description} \\
        \hline
        post\_project\_phases & Int Uniform & Min=2, Max=6 & 6 & Number of post-projection phases \\
        pre\_project\_phase\_len & Int Uniform & Min=2, Max=6 & Min=2, Max=7 & Length of pre-projection phase \\
        batch\_size & Fixed & 512 & 64 & Number of samples per training batch \\
        cluster\_coef & Normal & $\mu=-0.8,\sigma=0.5$ & $\mu=-0.8,\sigma=0.5$ & Coefficient of the cluster loss from \cite{chen2019looks} \\
        separation\_coef & Normal & $\mu=0.08,\sigma=0.1$ & $\mu=0.08,\sigma=0.1$ & Coefficient of the separation loss from \cite{chen2019looks} \\
        incorrect\_class\_connection & Fixed & 0 & -0.5 & Inital weight between prototypes and differing class logits \\
        joint\_lr\_step\_size & Int Uniform & Min=5, Max=5 & Min=2, Max=10 & Step size for LR schedule \\
        l1\_coef & Log Uniform & Min=$10^{-5}$, Max=0.001 & Min=$10^{-5}$, Max=0.001 & L1 regularization strength \\
        activation\_function & Fixed & cosine similarity & cosine similarity & Similarity function for prototype activation \\
        base\_latent\_dimension & Fixed & 2048 & 64 & Depth of the latent space before latent\_dim\_multiplier\_exp is applied \\
        latent\_dim\_multiplier\_exp & Int Uniform & Min=-3, Max=3 & Min=-3, Max=3 & Exponent controlling latent-dim scaling \\
        lr\_multiplier & Normal & $\mu=1,\sigma=0.4$ & $\mu=1,\sigma=0.4$ & Learning-rate scaling factor \\
        num\_addon\_layers & Int Uniform & Min=0, Max=2 & Min=0, Max=2 & Number of additional layers before prototype layer \\
        num\_prototypes\_per\_class & Fixed Values & 5 & 10, 20 & Number of prototypes allocated per class \\
        phase\_multipler & Fixed Values & 1, 2 & 1, 2 & Multiplier for training phase lengths \\
        position\_encoding\_strength & Fixed Values & 0.01, 0.05 & --- & Prototype feature position encoding strategy.  \\
        gen\_aug\_substitution\_rate & Fixed Values & 0, 0.025, 0.005, 0.01 & --- & Rate of nucleotide substitution for genetic augmentation \\
        gen\_aug\_insertion\_rate & Fixed Values & 0, 2, 4 & --- & Number of inserted for genetic augmentation \\
        gen\_aug\_deletion\_rate & Fixed Values & 0, 2, 4 & --- & Number of nucleotides deleted for genetic augmentation \\
        \hline
        \end{tabular}
        \caption{Hyperparameter search space for ProtoPNet.}
        \label{tab:sweep_tree}
    \end{table*}

    \begin{table*}[h!]
        \centering
        \footnotesize
        \begin{tabular}{l l l p{6cm}}
        \hline
        \textbf{Hyperparameter Name} & \textbf{Distribution} & \textbf{Values / Min-Max} & \textbf{Description} \\
        \hline
        modality\_reg\_coeff & Uniform & Min=-2, Max=1 & $\lambda_1$, the regularization coefficient for modality weights \\
        margin\_loss\_coeff & Uniform & Min=0, Max=0.5 & $\lambda_2$, the margin loss coefficient \\
        predictor\_mse\_coeff & Log Uniform & Min=$10^{-3}$, Max=1 & $\lambda_3$, the coefficient for the mean squared error loss of the predictor \\
        lr\_multiplier & Log Uniform & Min=0.1, Max=1 & Learning rate multiplier \\
        modality\_weight\_optimization\_epochs & Integer Uniform & Min=1, Max=6 & Number of epochs to optimize the modality weights and last layers \\
        ce\_alpha & Fixed value & 0.05 & Conformal confidence level used for margin and cross entropy loss \\
        batch\_size & Fixed value & 128 & Number of samples per training batch \\
        gen\_aug\_deletion\_rate & Integer Uniform & Min=0, Max=4 & Number of deletions applied during genetic data augmentation \\
        gen\_aug\_insertion\_rate & Integer Uniform & Min=0, Max=4 & Number of insertions applied during genetic data augmentation \\
        gen\_aug\_substitution\_rate & Fixed values & 0, 0.0001, 0.001, 0.005 & Substitution rates applied during genetic data augmentation \\
        \hline
        \end{tabular}
        \caption{Hyperparameter search space for Multimodal ProtoPNet.}
        \label{tab:sweep}
    \end{table*}

\subsubsection{Abstention Learning ProtoTree}

    Due to the hard-traversal of the ProtoTree, the number of leaf nodes must be greater than the number of classes (516) in order for the model to have enough capacity to route each class to its own leaf node. Therefore, the image, genetic, and multimodal ProtoTrees were all trained with a depth of 10, which consists of 1023 prototypes and $2^{10} = 1024$ leaf nodes. 
    
    The embedding dimension for the image and genetic models were $2048$ and $64$ respectively. We used cosine similarity as the similarity metric in both cases. 

    In a unimodal ProtoTree, the backbone, add-on layers, and the prototype layer are optimized with separate learning rates, while the prediction head is optimized through a gradient-free update rule. Training a ProtoTree consists of two stages. In the initial ``warm-up phase,'' we load and freeze the pretrained backbone and the prediction head, and we focus on training the add-on and prototype layers. In the ``joint phase,'' all layers are trained. We optimize the hyperparameters using a Bayesian optimization scheme that maximizes the best validation balanced accuracy. The descriptions and prior distributions of all hyperparameter values can be found in Table \ref{tab:sweep_tree}. 

    In our ALP, we initialize with the pretrained unimodal models and initialize both the modality weights and the logits in the leaves using threshold based modality assignment. We then optimize the hyperparameters in a similar fashion as done for the unimodal models, with descriptions and prior distributions of all hyperparameter values listed in Table \ref{tab:sweep_multimodal}. 

    \begin{table*}[h!]
        \centering
        \footnotesize
        \begin{tabular}{l l l l p{5cm}}
        \hline
        \textbf{Hyperparameter Name} & \textbf{Distribution} & \textbf{ProtoTree (Genetic)} & \textbf{ProtoTree (Image)} & \textbf{Description} \\
        \hline
        warm\_up\_phase\_len & Integer Uniform & Min=0, Max=50 & Min=0, Max=25 & The number of warm-up epochs before joint optimization begins \\
        joint\_phase\_len & Integer Uniform & Min=0, Max=50 & Min=0, Max=25 & The number of joint optimization epochs \\
        weight\_decay & Log Uniform & Min=$10^{-4}$, Max=1.0 & Min=0.1, Max=10.0 & Regularization parameter for weight decay \\
        num\_addon\_layers & Integer Uniform & Min=0, Max=3 & Min=0, Max=2 & The number of additional convolutional layers to add between the backbone and the prototype layer \\
        batch\_size & Fixed value & 1024 & 128 & Number of samples per training batch \\
        gen\_aug\_substitution\_rate & Uniform & Min=0.00, Max=0.01 & --- & Rate of nucleotide substitution for genetic data augmentation \\
        gen\_aug\_insertion\_rate & Integer Uniform & Min=0, Max=3 & --- & Rate of nucleotide insertion for genetic data augmentation \\
        gen\_aug\_deletion\_rate & Integer Uniform & Min=0, Max=3 & --- & Rate of nucleotide deletion for genetic data augmentation \\
        cluster\_coef & Uniform & Min=-1.5, Max=0.00 & Min=-1.5, Max=0.00 & Coefficient of the cluster loss from \cite{chen2019looks} \\
        orthogonality\_coef & Log Uniform & Min=$10^{-5}$, Max=$10^{-1}$ & Min=$10^{-5}$, Max=$10^{-1}$ & Coefficient of the orthogonality loss from \cite{wang2021interpretable} \\
        backbone\_lr & Log Uniform & Min=$10^{-5}$, Max=$10^{-2}$ & Min=$10^{-4}$, Max=$10^{-2}$ & Learning rate for the backbone network \\
        add\_on\_lr & Log Uniform & Min=$10^{-4}$, Max=$10^{-1}$ & Min=$10^{-4}$, Max=$10^{-2}$ & Learning rate for the add-on layers \\
        prototype\_lr & Log Uniform & Min=$10^{-3}$, Max=$10^{-1}$ & Min=$10^{-4}$, Max=$3\times10^{-2}$ & Learning rate for the prototype layer \\
        \hline
        \end{tabular}
        \caption{Hyperparameter search space for Image and Genetic ProtoTree.}
        \label{tab:sweep_tree}
    \end{table*}
    
    \begin{table*}[h!]
        \centering
        \footnotesize
        \begin{tabular}{l l l p{5cm}}
        \hline
        \textbf{Hyperparameter Name} & \textbf{Distribution} & \textbf{ProtoTree (Multimodal)} & \textbf{Description} \\
        \hline
        warm\_up\_phase\_len & Integer Uniform & Min=0, Max=50 & The number of warm-up epochs before joint optimization begins \\
        joint\_phase\_len & Integer Uniform & Min=0, Max=50 & The number of joint optimization epochs \\
        batch\_size & Fixed value & 128 & Number of samples per training batch \\
        gen\_aug\_substitution\_rate & Uniform & Min=0.00, Max=0.01 & Rate of nucleotide substitution for genetic data augmentation \\
        gen\_aug\_insertion\_rate & Integer Uniform & Min=0, Max=3 & Rate of nucleotide insertion for genetic data augmentation \\
        gen\_aug\_deletion\_rate & Integer Uniform & Min=0, Max=3 & Rate of nucleotide deletion for genetic data augmentation \\
        cluster\_coef & Uniform & Min=-1.5, Max=0.00 & The value of $\lambda_{clst}$ for clustering loss from \cite{chen2019looks} \\
        orthogonality\_coef & Log Uniform & Min=$10^{-5}$, Max=0.1 & The value of $\lambda_{2}$ for orthogonality loss from \cite{wang2021interpretable} \\
        variability\_coef & Uniform & Min=-1.5, Max=0.00 & The value of $\lambda_{3}$ for variability loss \\
        weight\_decay & Log Uniform & Min=$10^{-5}$, Max=1.0 & The value of $\lambda_4$ for weight decay \\
        routing\_coef & Fixed & 0.01 & The value of $\lambda_{5}$ for routing loss \\
        backbone\_lr & Log Uniform & Min=$10^{-5}$, Max=0.01 & Learning rate for the backbone network \\
        prototype\_lr & Log Uniform & Min=0.001, Max=0.1 & Learning rate for the prototype layer \\
        add\_on\_lr & Log Uniform & Min=$10^{-4}$, Max=0.1 & Learning rate for the add-on layers \\
        \hline
        \end{tabular}
        \caption{Hyperparameter search space for Multimodal ProtoTree.}
        \label{tab:sweep_multimodal}
    \end{table*}
    
\subsection{Hardware and Software}

    Each experiment was conducted on a server consisting of 4 Intel Xeon E5-2640v4 CPU cores and a single RTX A5000 GPU with 24GB of memory. All models were run with PyTorch v1.13.1 and CUDA version 11.3.  

\section{Local Analysis}

    In this section, we visualize the most similar prototypes to a given image/genetic input pair. We call this the \textit{local analysis} of the input, as in \cite{chen2019looks}. We provide local analyses for a few samples from BIOSCAN-1M on both CAL and ALP. 

\subsection{Conformal Abstention Learning}

    \begin{figure*}
        \centering
        \includegraphics[width=0.6\linewidth]{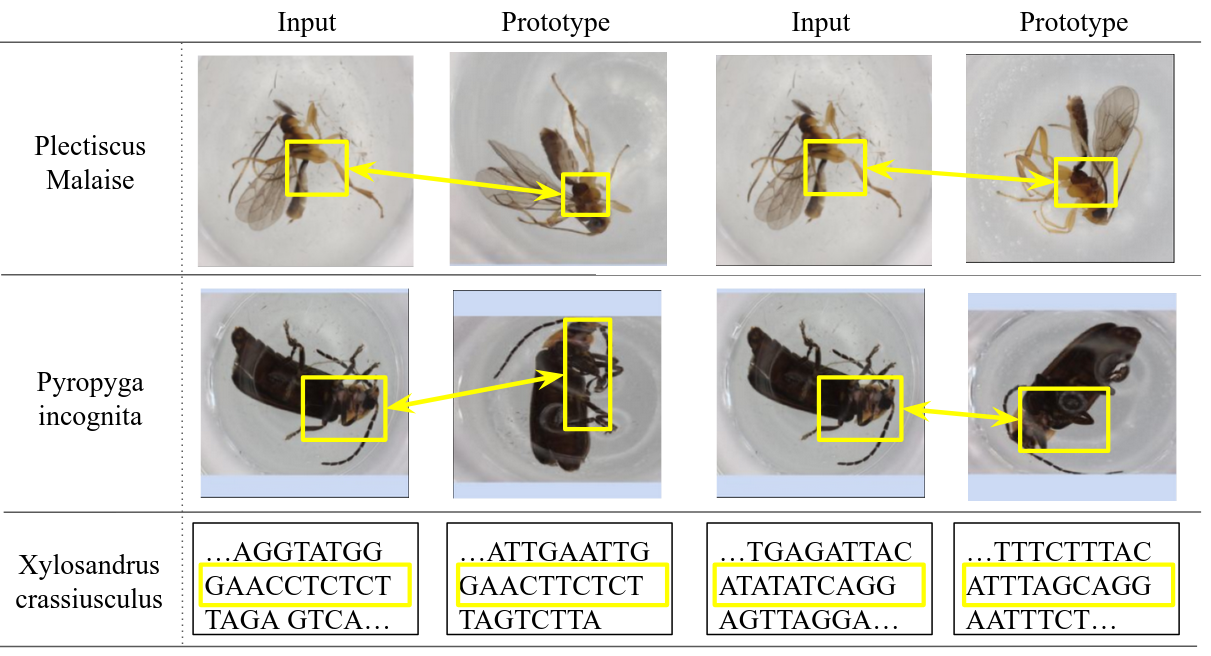}
        \caption{Local analyses of three test samples from BIOSCAN-1M for a CAL ProtoPNet. For each sample, we show the two most similar image or genetic prototypes in our model.}
        \label{fig:local_analysis_cal.png}
    \end{figure*}

    Figure \ref{fig:local_analysis_cal.png} shows the closest prototypes to two image samples and one genetic sample from a CAL ProtoPNet. Unsurprisingly, the most similar image prototypes come from the same species as the input. For sample $1$, both of the nearest prototypes correspond to the abdomen of the \textit{Pletiscus Malaise}. For sample $2$, both prototypes correspond to the upper legs of the \textit{Pyropyga Incognita}. Both of the nearest prototypes to the genetic sample show a perfect match (activation of $1$). 

\subsection{Abstention Learning ProtoTree}

\begin{figure*}
        \centering
        \includegraphics[width=0.6\linewidth]{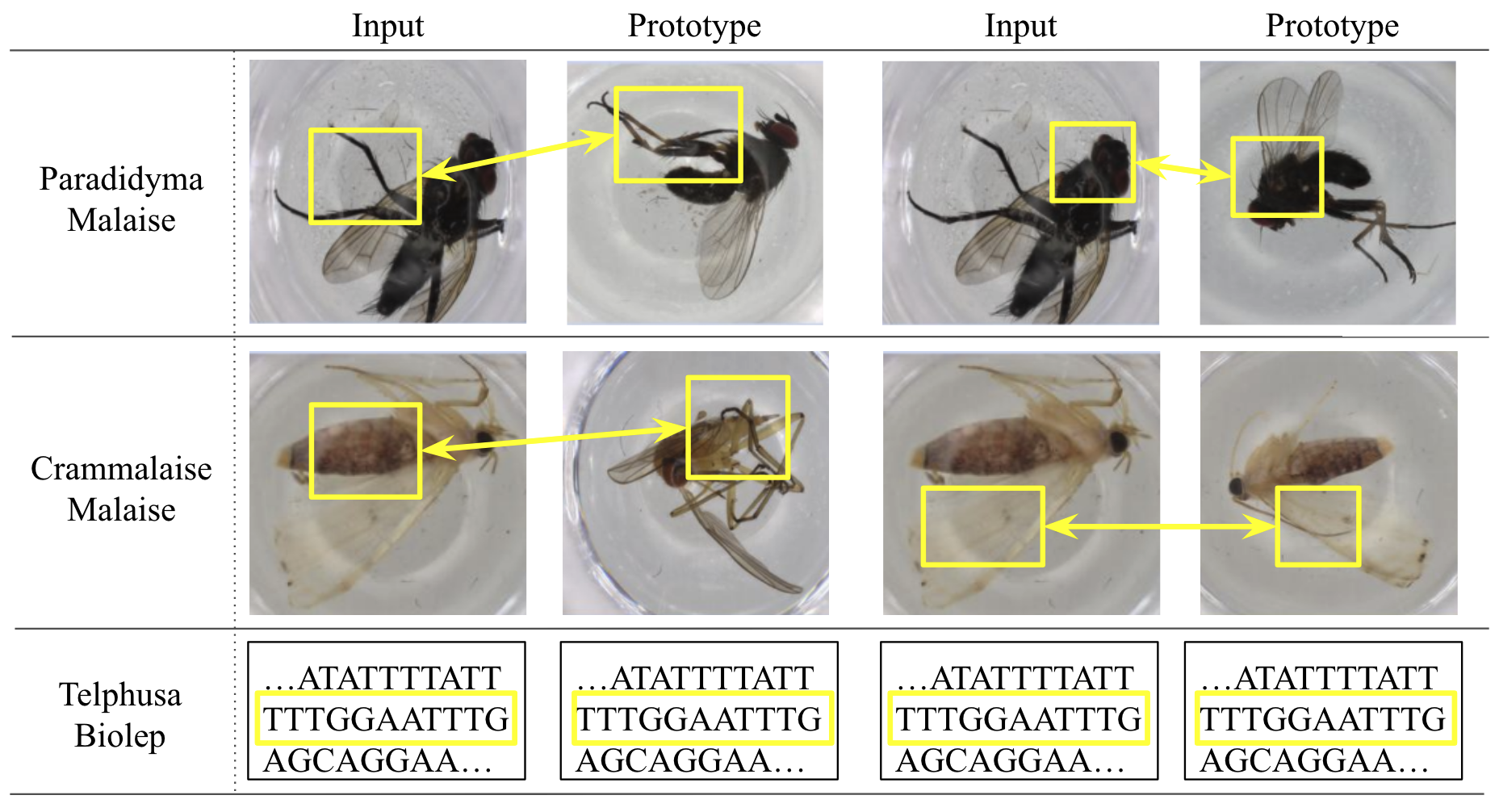}
        \caption{Local analyses of three test samples from BIOSCAN-1M for an abstention learning ProtoTree. For each sample, we show the two most similar image or genetic prototypes in our model. }
        \label{fig:local_analysis_prototree.png}
    \end{figure*}

    Figure \ref{fig:local_analysis_prototree.png} shows the local analyses of two image samples and one genetic sample from BIOSCAN-1M, where the prototypes are taken from a fully trained ALP with threshold $t = 0.8$. In the figure, we observe semantic similarities between certain patches in the input image and the image prototypes such as the presence of spiny legs or a black hairy head in the species \textit{Paradidyma Malaise}. For \textit{Crammalaise Malaise}, the model identifies prototypes indicating a brown ellipsoid body and large triangular wings. Again, the most similar image prototypes also come from samples that are of the same species as the input. The local analysis on the genetic input sequence shows a perfect match (activation of $1$) with prototypes that have been projected to other samples of the same species. 

\section{Global Analysis}

    In this subsection, we visualize the most similar image and genetic samples to a given image or genetic prototype. We refer to this as \textit{global analysis}, as in \cite{chen2019looks}. We provide global analyses for a few samples from BIOSCAN-1M on both CAL and ALP. 

\subsection{Conformal Abstention Learning}

\begin{figure*}
        \centering
        \includegraphics[width=0.8\linewidth]{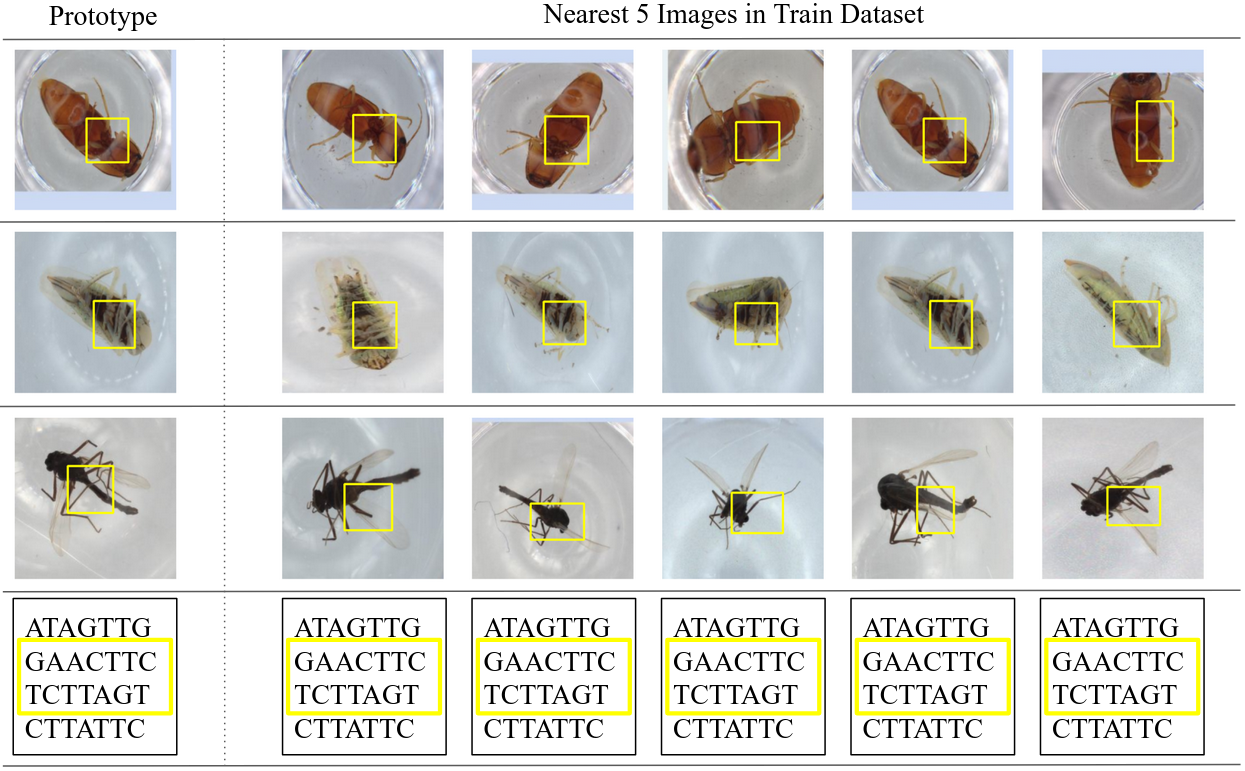}
        \caption{Global analyses of three image prototypes and one genetic prototype from BIOSCAN-1M for a CAL ProtoPNet. For each prototype, we show the 5 most similar images to the prototype. }
        \label{fig:global_analysis_protopnet.png}
    \end{figure*}

    Figure \ref{fig:global_analysis_protopnet.png} presents 4 prototypes---three image and one genetic---and shows the top 5 most similar samples to each of the prototypes learned by the CAL ProtoPNet. The prototypes visually match the corresponding patches of highest activation in the 5 samples. Furthermore, the closest genetic samples all match perfectly to the prototype with an activation of $1$. Often, the most similar samples for a given prototype come from the same class as that of the prototype.

\subsection{Abstention Learning ProtoTree}

\begin{figure*}
        \centering
        \includegraphics[width=0.8\linewidth]{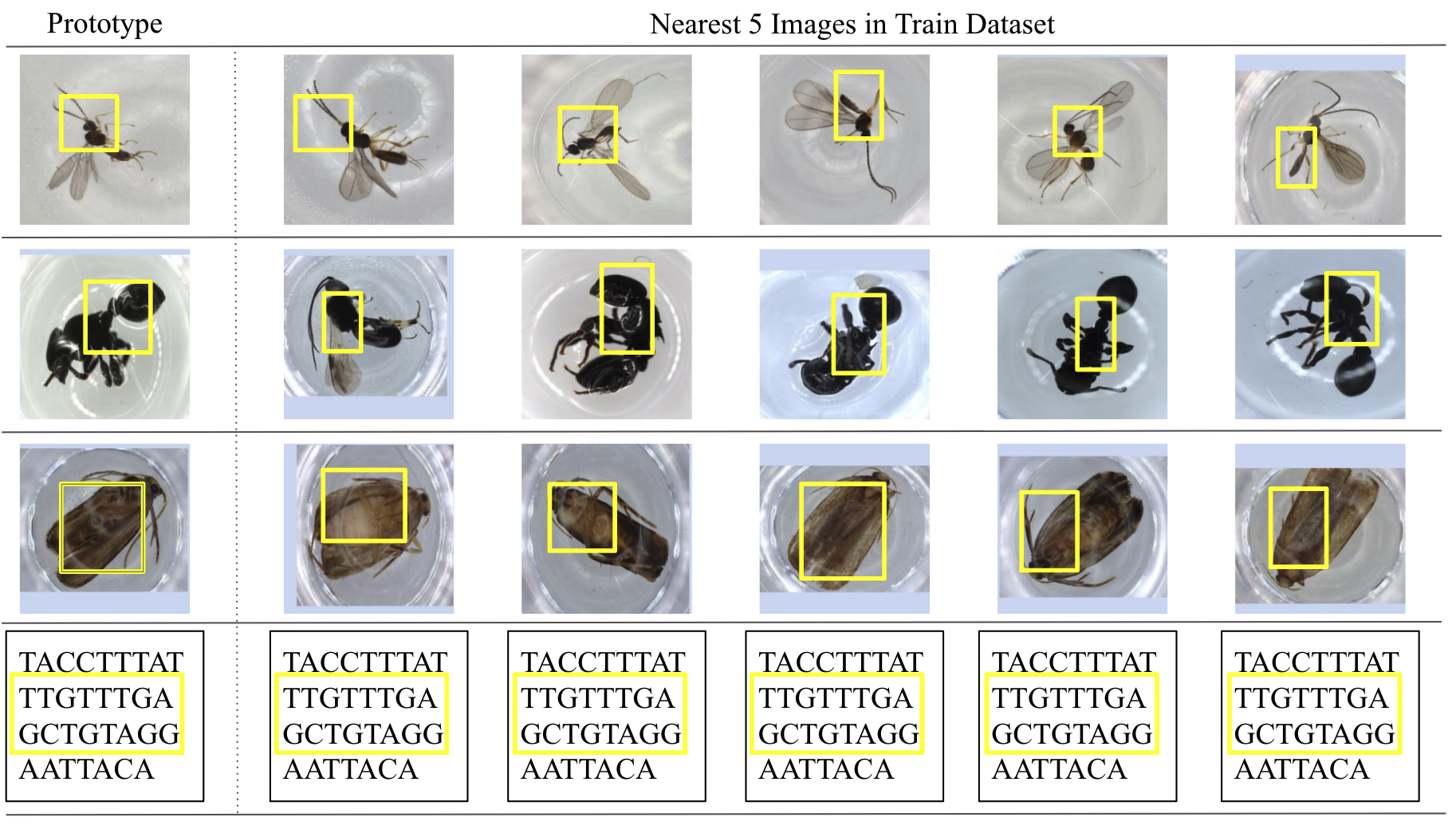}
        \caption{Global analyses of three image prototypes and one genetic prototype from BIOSCAN-1M for an abstention learning ProtoTree. For each prototype, we show the 5 most similar images to the prototype. We can describe the prototypes as such: a tiny head with long antennae (1st row), a black skinny thorax (2nd row), large brown rectangular wings (3rd row), and a genetic sequence (4th row).}
        \label{fig:global_analysis_prototree.png}
    \end{figure*}

    Figure \ref{fig:global_analysis_prototree.png} presents 4 prototypes---three image and one genetic. It shows the top 5 most similar samples to each of the prototypes learned by the ALP, which was trained with threshold $t = 0.8$. Once again, each image prototype semantically matches the corresponding patches of highest activation in the top 5 samples, and each genetic prototype has a perfect match with corresponding genetic patches.  

\end{document}